\documentclass[runningheads]{ECSQARUfiles/llncs}

\usepackage{amsmath,amsfonts,bm}

\def\eqref#1{equation~\ref{#1}}
\def\1{\bm{1}}

\DeclareMathAlphabet{\mathsfit}{\encodingdefault}{\sfdefault}{m}{sl}
\SetMathAlphabet{\mathsfit}{bold}{\encodingdefault}{\sfdefault}{bx}{n}

\DeclareMathOperator*{\argmin}{arg\,min}

\usepackage{hyperref}
\usepackage{url}

\newcommand{\norm}[1]{\left\|#1\right\|}

\usepackage[vlined,ruled]{algorithm2e}

\usepackage[utf8]{inputenc} % allow utf-8 input
\usepackage[T1]{fontenc}    % use 8-bit T1 fonts
\usepackage{hyperref}       % hyperlinks
\usepackage{url}            % simple URL typesetting
\usepackage{booktabs}       % professional-quality tables
\usepackage{amsfonts}       % blackboard math symbols
\usepackage{nicefrac}       % compact symbols for 1/2, etc.
\usepackage{microtype}      % microtypography
\usepackage{thm-restate}

\usepackage{graphicx}
\usepackage{subfigure}
\usepackage{changes}

\usepackage{amsmath}
\usepackage{multirow}
\usepackage{hhline}
\usepackage{wrapfig}
\usepackage{caption}
\usepackage{enumitem}
\usepackage[export]{adjustbox}
\usepackage{amsmath,mathtools}

\makeatletter
\newcommand{\raisemath}[1]{\mathpalette{\raisem@th{#1}}}
\newcommand{\raisem@th}[3]{\raisebox{#1}{$#2#3$}}
\makeatother

\DeclarePairedDelimiter\abs{\lvert}{\rvert}%

\newcommand{\uglad}{{\texttt{uGLAD}~}}

\newcommand{\ngm}{{\texttt{NGM~}}}
\newcommand{\ngms}{{\texttt{NGMs~}}}
\newcommand{\ngmns}{{\texttt{NGM}}}
\newcommand{\ngmsns}{{\texttt{NGMs}}}

\usepackage[T1]{fontenc}

\begin{document}

\title{Neural Graphical Models}

\author{Harsh Shrivastava \&
Urszula Chajewska} 
\institute{Microsoft Research, Redmond, USA\\Contact:\{hshrivastava,urszc\}@microsoft.com}
% \email{\{hshrivastava,urszc\}@microsoft.com}

\titlerunning{Neural Graphical Models}
\authorrunning{H. Shrivastava \and U. Chajewska}

\maketitle

\vskip 0.3in

\newcommand{\fix}{\marginpar{FIX}}
\newcommand{\new}{\marginpar{NEW}}

\newcommand{\Rho}{\mathrm{P}}

\begin{abstract}

Probabilistic Graphical Models are often used to understand dynamics of a system. They can model relationships between features (nodes) and the underlying distribution.  Theoretically these models can represent very complex dependency functions, but in practice often simplifying assumptions are made due to computational limitations associated with graph operations. In this work we introduce Neural Graphical Models (\ngmsns) which attempt to represent complex feature dependencies with reasonable computational costs. Given a graph of feature relationships and corresponding samples, we capture the dependency structure between the features along with their complex function representations by using a neural network as a multi-task learning framework. We provide efficient learning, inference and sampling algorithms. \ngms can fit generic graph structures including directed, undirected and mixed-edge graphs as well as support mixed input data types. We present empirical studies that show \ngmsns' capability to represent Gaussian graphical models, perform inference analysis of a lung cancer data and extract insights from a real world infant mortality data provided by CDC.

\textit{Software}:{\small\url{https://github.com/harshs27/neural-graphical-models}}
% {\small\href{https://github.com/harshs27/neural-graphical-models}{~\ngm code link}}

\end{abstract}


\section{Introduction}

Graphical models are a powerful tool to analyze data. They can represent the relationships between features and provide underlying distributions that model functional dependencies between them~\cite{pearl88,koller2009probabilistic}. 
Learning, inference and sampling are operations that make such graphical models useful for domain exploration. Learning, in a broad sense, consists of fitting the distribution function parameters from data. Inference is the procedure of answering queries in the form of conditional distributions with one or more observed variables. Sampling is the ability to draw samples from the underlying distribution defined by the graphical model. One of the common bottlenecks of graphical model representations is having high computational complexities for one or more of these procedures. 

In particular, various graphical models have placed restrictions on the set of distributions or types of variables in the domain. Some graphical models work with continuous variables only (or categorical variables only) or place restrictions on the graph structure (e.g., that continuous variables cannot be parents of categorical variables in a DAG).  Other restrictions affect the set of distributions the models are capable of representing, e.g., to multivariate Gaussian. 

For wide adoption of graphical models, the following properties are desired: 

\begin{itemize}[leftmargin=*,nolistsep]
    \item Rich representations of complex underlying distributions.
    \item Ability to simultaneously handle various input types such as categorical, continuous, images and embedding representations. 
    \item Efficient algorithms for learning, inference and sampling. 
    \item Support for various representations: directed, undirected, mixed-edge graphs.
    \item Access to the learned underlying distributions for analysis. 

\end{itemize}
In this work, we propose 
Neural Graphical Models (\ngmsns) that satisfy the aforementioned desiderata in a computationally efficient way.  
\ngms accept a feature dependency structure that can be given by an expert or learned from data.  The dependency structure may have the form of a graph with clearly defined semantics (e.g., a Bayesian network graph or a Markov network graph) or an adjacency matrix. Note that the graph may be either directed or undirected.  Based on this dependency structure, \ngms learn to represent the probability function over the domain using a deep neural network.  The parameterization of such a network can be learned from data efficiently, with a loss function that jointly optimizes adherence to the given dependency structure and fit to the data. Probability functions represented by \ngms are unrestricted by any of the common restrictions inherent in other PGMs.  They also support efficient inference and sampling.

\section{Related works}
\label{sec:related}

Probabilistic Graphical Models (PGMs) aim to learn the underlying joint distribution from which input data is 
sampled. Often, to make learning of the distribution computationally feasible, inducing an independence graph structure between the features helps. In cases where this independence graph structure is provided by a domain expert, the problem of fitting PGMs reduces to learning distributions over this graph. Alternatively, there are many methods traditionally used to jointly learn the structure as well as the parameters~\cite{heckerman1995learning,spirtes1995learning,koller2009probabilistic,scanagatta2019survey} and have been widely used to analyse data in many domains~\cite{barton2012bayesian,bielza2014bayesian,borunda2016bayesian,shrivastava2019cooperative,shrivastava2020using,aluru2021engrain}. 

Recently, many interesting deep learning based approaches for DAG recovery have been proposed~\cite{zheng2018dags,zheng2020learning,lachapelle2019gradient,yu2019dag}. These works primarily focus on structure learning but technically they are learning a Probabilistic Graphical Model. These works depend on the existing algorithms developed for the Bayesian networks for the inference and sampling tasks. A parallel line of work combining graphical models with deep learning are Bayesian deep learning approaches: Variational AutoEncoders, Boltzmann Machines etc.~\cite{li2016combining,johnson2016composing,wang2020survey}. The deep learning models have significantly more parameters than traditional Bayesian networks, which makes them less suitable for datasets with a small number of samples. Using these deep graphical models for downstream tasks is computationally expensive and often impedes their adoption.

We would be remiss not to mention the technical similarities \ngms have with some recent research works. 
We found "Learning sparse nonparametric DAGs"~\cite{zheng2020learning} to be the closest in terms of representation ability. In one of their versions, they model each independence structure with a different neural network (MLP). However, their choice of modeling feature independence criterion differs from \ngmns. They zero out the weights of the row in the first layer of the neural network to induce independence between the input and output features. This 
formulation restricts them from sharing the NNs across different factors. Second, we found in~\cite{lachapelle2019gradient} path norm formulations of using the product of NN weights for input to output connectivity similar to \ngmsns. They used the path norm to parametrize the DAG constraint for continuous optimization, while~\cite{shrivastava2020grnular,shrivastava2022grnular} used them within unrolled algorithm framework to learn sparse gene regulatory networks. 

Methods that model the conditional independence graphs~\cite{friedman2008sparse,belilovsky2017learning,shrivastava2019glad,shrivastava2022uglad,shrivastava2022a,shrivastava2022methods} are a type of graphical models that are based on underlying multivariate Gaussian distribution. Probabilistic Circuits~\cite{peharz2020einsum}, Conditional Random Fields or Markov Networks~\cite{sutton2012introduction} and some other PGM formulations like~\cite{uria2013rnade,uria2016neural,yang2014mixed,molina2018mixed} are popular. These PGMs often make simplifying assumptions about the underlying distributions and place restrictions on the accepted input data types.  
Real-world input data often consist of mixed datatypes (real, categorical, text, images etc.) and it is  challenging for the existing graphical model formulations to handle.

\section{Neural Graphical Models}
\label{sec:ngm}
We propose a new Probabilistic Graphical Model type, called Neural Graphical Models (\ngmsns) and describe the associated learning, inference and sampling algorithms. Our model accepts all input types and avoids placing any restrictions on the form of underlying distributions. 

\textbf{Problem setting}: 
We consider input data \textbf{X} that have $M$ samples with each sample consisting of $D$ features. An example can be gene expression data, where we have a matrix of the microarray expression values (samples) and genes (features). In the medical domain, we can have a mix of continuous and categorical data describing a patient's health. We are also provided a graph \textbf{G} which can be directed, undirected or have mixed-edge types that represents our belief about the feature dependency relationships (in a probabilistic sense). Such graphs are often provided by experts and include inductive biases and domain knowledge about the underlying system functions. In cases where the graph is not provided, we make use of the state-of-the-art algorithms to recover DAGs or CI graphs, refer to Sec.~\ref{sec:related}. 
The \ngm input is the tuple (\textbf{X}, \textbf{G}).

\begin{figure}%[b]
\centering 
\subfigure{\label{fig:r1c1}\includegraphics[width=65mm, height=35mm]{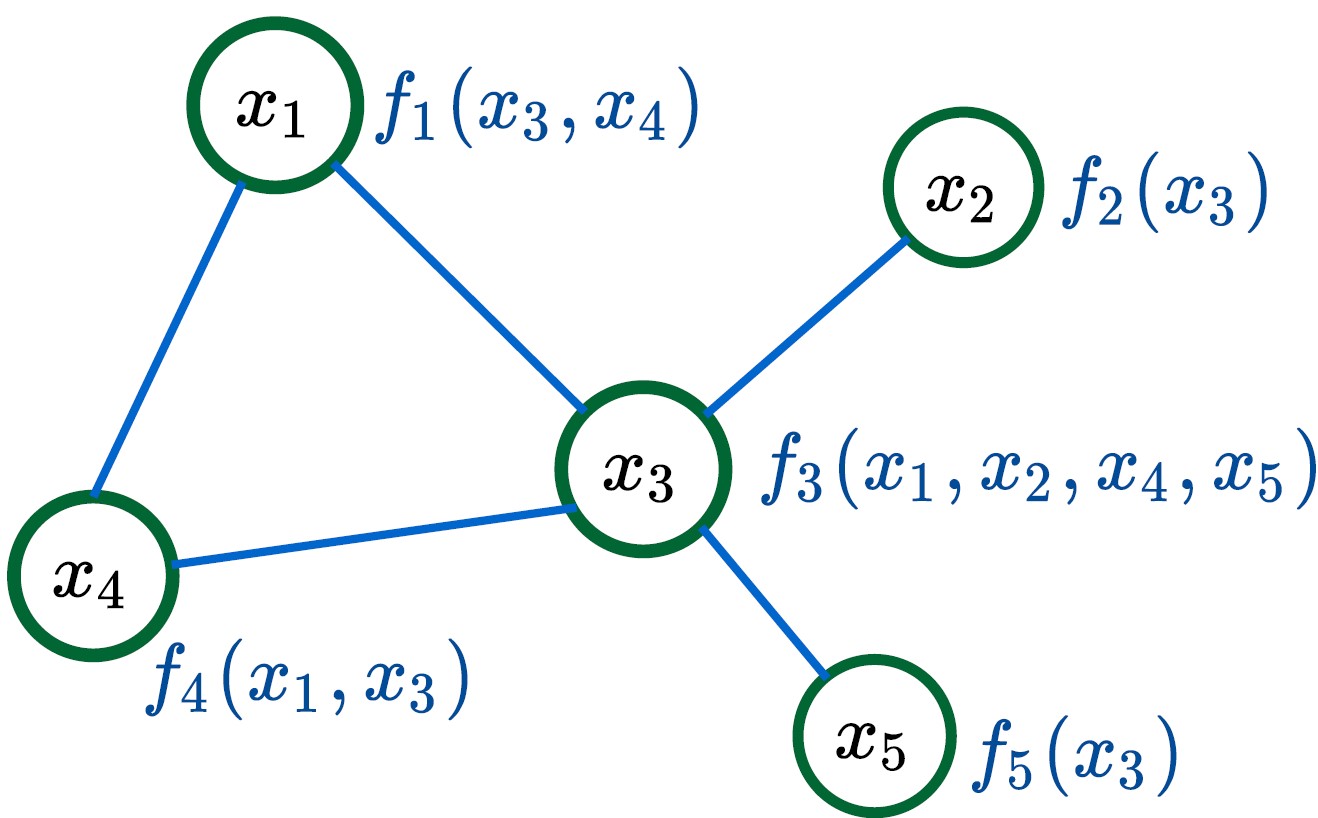}}\qquad
\subfigure{\label{fig:r1c2}\includegraphics[width=40mm]{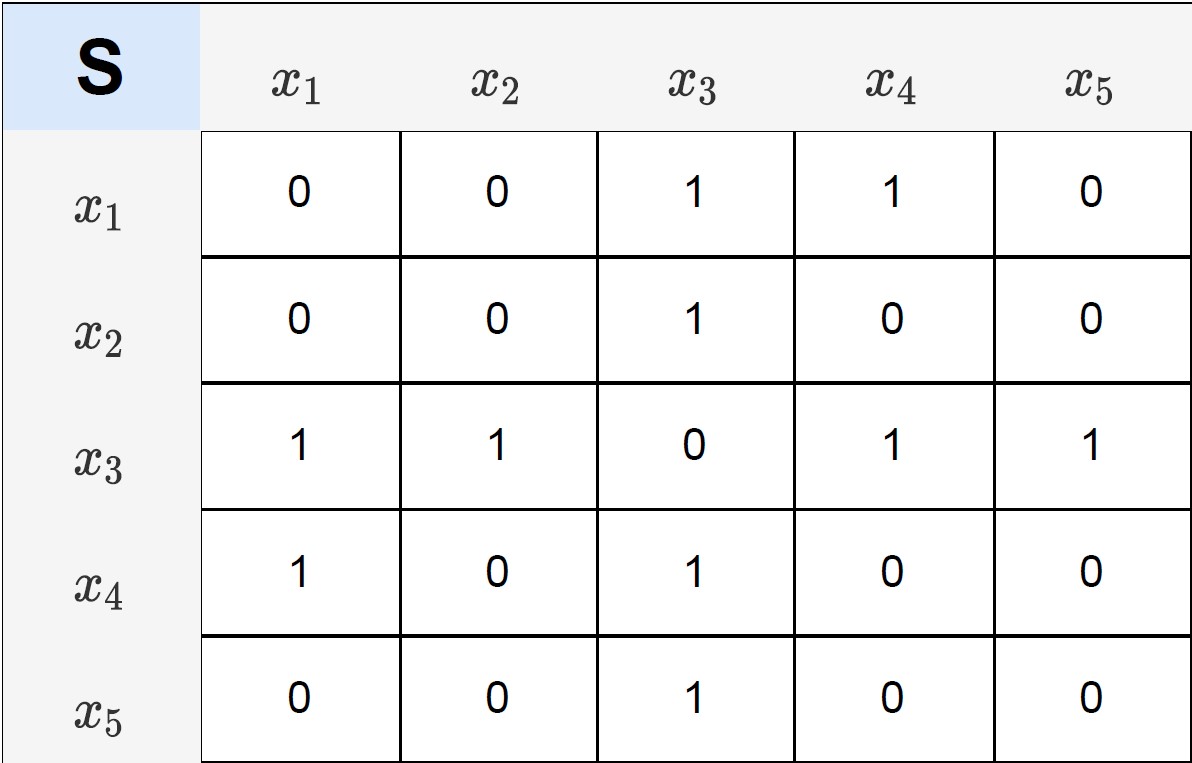}}
\caption{\small \textbf{Graphical view of \ngmsns}: The input graph \textbf{G} (undirected) for given input data $X\in\mathbb{R}^{M\times D}$. Each feature $x_i = f_i(\text{Nbrs}(x_i))$ is a function of the neighboring features. For a DAG, 
the functions between features will be defined by the Markov Blanket relationship $x_i = f_i(\text{MB}(x_i))$. The adjacency matrix (right) represents the associated dependency structures.}
\label{fig:graphical-view}
\vspace{-5mm}
\end{figure}

\subsection{Representation}
Fig.~\ref{fig:graphical-view} shows a sample graph recovered and how we view the value of each feature as a function of the values of its neighbors. 
For directed graphs, each feature's value is represented as a function of its Markov blanket in the graph.
We use the graph \textbf{G} to understand the domain's dependency structure, but ignore any potential parametrization associated with it.

\begin{figure}%[b]
\centering 
\includegraphics[width=80mm]{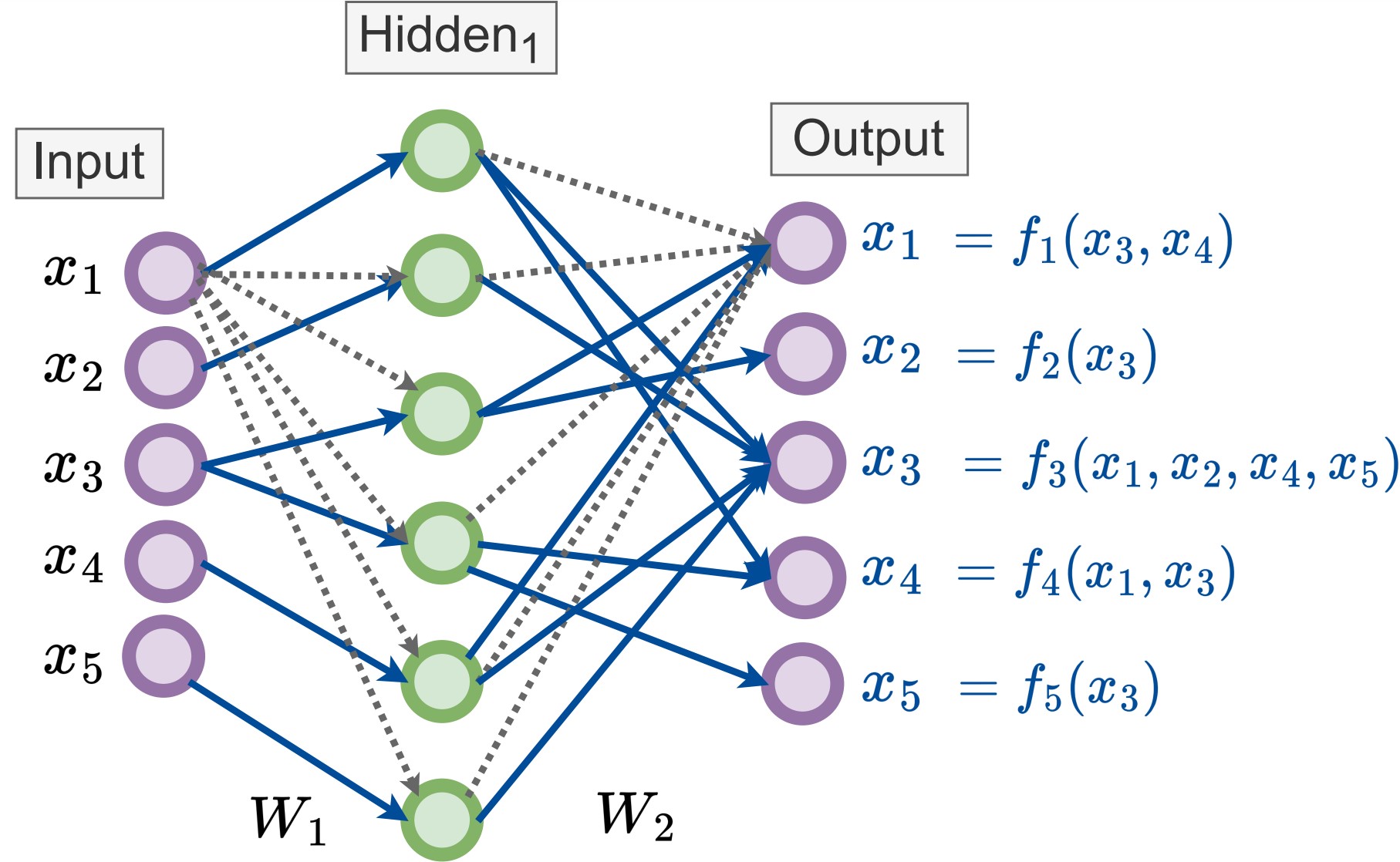}
\caption{\small \textbf{Neural view of \ngmsns}: NN as a multitask learning architecture capturing non-linear dependencies for the features of the undirected graph in Fig.~\ref{fig:graphical-view}. If there is a path from the input feature to an output feature, that indicates a dependency between them. The dependency matrix between the input and output of the NN reduces to matrix product operation $S_{nn} = \Pi_i |W_i|=|W_1|\times |W_2|$. 
Note that not all the zeroed out weights of the MLP (in black-dashed lines) are shown for the sake of clarity.}
\label{fig:neural-view}
\vspace{-4mm}
\end{figure}

We introduce a \textit{neural view} which is another way of representing \textbf{G}, as shown in Fig.~\ref{fig:neural-view}. The neural networks used are multi-layer perceptrons with appropriate input and output dimensions 
that represent graph connections in \ngmsns. 
%Specifically, we view the NNs as a `white-box' and focus on the paths from input to output that represent functional dependencies. 
We denote a NN with $L$ number of layers with the weights $\mathcal{W}=\{W_1, W_2, \cdots, W_L\}$ and biases $\mathcal{B}=\{b_1, b_2, \cdots, b_L\}$ as $f_{\mathcal{W, B}}(\cdot)$ with non-linearity not mentioned explicitly. We experimented with multiple non-linearities and found that $\operatorname{ReLU}$ fits well with our framework. Applying the NN to the input $X$ evaluates the following mathematical expression, $f_{\mathcal{W, B}}(X) =\operatorname{ReLU}(W_L\cdot(\cdots(W_2\cdot\operatorname{ReLU}(W_1\cdot X + b_1) + b_2)\cdots)+b_L) 
$.
The dimensions of the weights and biases are chosen such that the neural network input and output units are equal to $|\mathcal{D}|$ with the hidden layers dimension $H$ remaining a design choice. In experiments, we start with $H=2|\mathcal{D}|$ and subsequently adjust the dimensions based on the validation loss. The product of the weights of the neural networks $S_{nn} = { \prod_{l=1}^L }\abs{W_l}=|W_1|\times |W_2|\times \cdots \times |W_L|$, where $|W|$ computes the absolute value of each element in $W$, gives us path dependencies between the input and the output units. For short hand, we denote $S_{nn}=\Pi_i|W_i|$. If $S_{nn}[x_i,x_o]=0$, then the output unit $x_o$ is independent of the input unit $x_i$. Increasing the layers and hidden dimensions of the NNs provide us with richer dependence function complexities.

Initially, the NN is fully connected.  Some of the connections will be dropped during training, as the associated weights are zeroed out.  We can view the resulting NN as a \textit{glass-box} model (indicating transparency), since we can discover functional dependencies by analyzing paths from input to output.

\subsection{Learning}\label{sec:ngm-learning}

Using the rich and compact functional representation achieved by using the \textit{neural} view, the learning task is to fit the neural networks to achieve the desired dependency structure \textbf{S} (encoded by the input graph \textbf{G}), along with fitting the regression to the input data \textbf{X}. 
Given the input data \textbf{X} 
we want to learn the functions as described by the \ngms \textit{graphical} view, Fig.~\ref{fig:graphical-view}. These can be obtained by solving the multiple regression problems shown in neural view, Fig.~\ref{fig:neural-view}. We achieve this by using the neural view as a multi-task learning framework. The goal is to find the set of parameters $\mathcal{W}$ that minimize the loss expressed as the distance from $X^k$ to $f_{\mathcal{W}}(X^k)$ (averaged over all samples $k$) while maintaining the dependency structure provided in the input graph \textbf{G}. We can define the regression operation as follows:
\begin{align}\label{eqn:learning-regression}
    \argmin_{\mathcal{W,B}}& \sum_{k=1}^{M} \norm{X^k - f_{\mathcal{W,B}}(X^k)}^2_2 \quad  %\\\nonumber
    %&
    s.t. \left(\Pi_{i=1}^L |W_i|\right) * S^c = 0
\end{align}
where we introduced a \textit{soft-graph} constraint. Here, $S^c$ represents the complement of the matrix $S$, which essentially replaces $0$ by $1$ and vice-versa. The $A*B$ represents the Hadamard operator which does an element-wise matrix multiplication between the same dimension matrices $A, B$. Including the constraint as a Lagrangian term with $\ell_1$ penalty and a constant $\lambda$ that acts a tradeoff between fitting the regression and matching the graph dependency structure, we get the following optimization formulation
\begin{align}\label{eqn:optimization-function}
    \argmin_{\mathcal{W,B}} \sum_{k=1}^{M} \norm{X^k - f_{\mathcal{W,B}}(X^k)}^2_2 %\\ \nonumber
    + \lambda \log{\left(\norm{\left(\Pi_{i=1}^{L} |W_i|\right) * S^c}_1\right)}
\end{align}

In our implementation, the individual weights are normalized using $\ell_2$-norm before taking the product. We normalize the regression loss and the structure loss terms and apply appropriate scaling to the input data features.
 
\textbf{Proximal Initialization strategy}:
To get a good initialization for the NN parameters $\mathcal{W}$ and $\lambda$ we implement the following procedure. We solve the regression problem described in Eq.~\ref{eqn:learning-regression} without the structure constraint. This gives us a good initial guess of the NN weights $\mathcal{W}^0$. 
We choose the value $\lambda=\norm{\left(\Pi_i |W^0_i|\right) * S^c}^2_2$ and update after each epoch. Experimentally, we found that this strategy may not work optimally in a few cases and in such cases we recommend fixing the value of $\lambda$ at the beginning of the optimization. The value of $\lambda$ can be chosen such that it brings the regression loss and the structure loss values to same scale.

\begin{wrapfigure}[23]{R}{0.68\textwidth}
\vspace{-9mm}
\begin{algorithm}[H]
\caption{\ngmsns: Learning algorithm}
\label{algo:ngms-learning}
  \DontPrintSemicolon
  %\SetKwFunction{Grad}{Grad}
  \SetKwProg{Fn}{Function}{:}{}
  \SetKwFor{uFor}{For}{do}{}
  \SetKwFor{ForPar}{For all}{do in parallel}{}
  \SetKwFunction{fitngm}{fit-ngm}
  \SetKwFunction{ngmlearn}{ngm-learning}
  \SetKwFunction{proxinit}{proximal-init}
%   \vspace{2mm}
    \Fn{\proxinit{$X, S$}}{
      $f_{\mathcal{W}}\gets$ Init MLP using dimensions from S\; %\\
      $f_{\mathcal{W}^0} \leftarrow \argmin_{\mathcal{W}} \sum_{k=1}^{M} \norm{X^k - f_{\mathcal{W}}(X^k)}^2_2$\; %\\
      (Using Adam optimizer for $E_1$ epochs)\; %\\
     \KwRet $f_{\mathcal{W}^0}$ 
    }
    \vspace{1mm}
    \Fn{\fitngm{$X, S, f_{\mathcal{W}^0}, \lambda^0$}}{
      \uFor{$e = 1,\cdots, E_2$}{
        % $f_{\mathcal{W}^e} \gets \argmin_{\mathcal{W}} \sum_{k=1}^{M} \norm{X_{\mathcal{D}}^k - f_{\mathcal{W}}(X_\mathcal{O}^k)}^2$ \;\qquad\qquad\qquad$+ \lambda^{e-1} \norm{\left(\Pi_i |W_i^{e-1}|\right) * S^c}_1$\;
        $\mathcal{L}_\text{Lr}$ = $ \sum_{k=1}^{M} \norm{X^k - f_{\mathcal{W}^{e-1}}(X^k)}^2_2$ \; %\\ 
        $~~~~~~~~+ \lambda^{e-1} \log{\norm{\left(\Pi_i |W_i^{e-1}|\right) * S^c}_1}$\; %\\
        $\mathcal{W}^{e} \gets $ backprop $\mathcal{L}_\text{Lr}$ to update params\; %\\
        $\cdots$ (optional $\lambda$ update) $\cdots$\; %\\
        $\lambda^{e} \gets \norm{\left(\Pi_i |W^{e}_i|\right) * S^c}^2_2$ \; %\\
        % Detach $\lambda^{e}$ from the computational graph\; 
        }
     \KwRet $f_\mathcal{W}$%$\Theta,Z,\lambda$
    }
    \vspace{1mm}
    \Fn{\ngmlearn{$X, S$}}{
        $f_{\mathcal{W}^0} \gets $ \proxinit{$X, S$}\;% \\
        $\lambda^0\gets\norm{\left(\Pi_i |W^0_i|\right) * S^c}^2_2$\; %\\
        $f_{\mathcal{W}} \gets$\fitngm{$X, S, f_{\mathcal{W}^0}, \lambda^0$}\; %\\
        \KwRet $f_{\mathcal{W}}$
    }%\vspace{-mm}
\end{algorithm}
\end{wrapfigure}

The learned \ngm describes the underlying graphical model distributions, as presented in Alg.~\ref{algo:ngms-learning}. There are multiple \textbf{benefits} of jointly optimizing in a multi-task learning framework modeled by the neural view of \ngmsns, Eq.~\ref{eqn:optimization-function}. First, sharing of parameters across tasks helps in significantly reducing the number of learning parameters. It also makes the regression task more robust with respect to noisy and anomalous data points. A separate regression model for each feature may lead to inconsistencies in the learned distribution~\cite{DependencyNetworks}. Second, we fully leverage the expressive power of the neural networks to model complex non-linear dependencies. 
Additionally, learning all the functional dependencies jointly allows us to leverage batch learning powered with GPU based scaling to get quicker runtimes.

\subsection{Extension to Generic Data Types}

%The learning, inference and sampling algorithms proposed for \ngms can be extended to any generic input data type. 
In real world applications, we often find inputs consisting of generic datatypes. For instance, in the gene expression data, there can be meta information (categorical) or images associated with the genes. Optionally, including  node embeddings from pretrained deep learning models can be useful. These variables are dependent on each other and can be represented in the form of a graph that acts as an input to \ngmns. We present two approaches for \ngms to handle such \textbf{mixed} input data types simultaneously which are otherwise difficult to accommodate in the existing PGM frameworks.

\begin{figure*}
\centering 
\includegraphics[width=120mm]{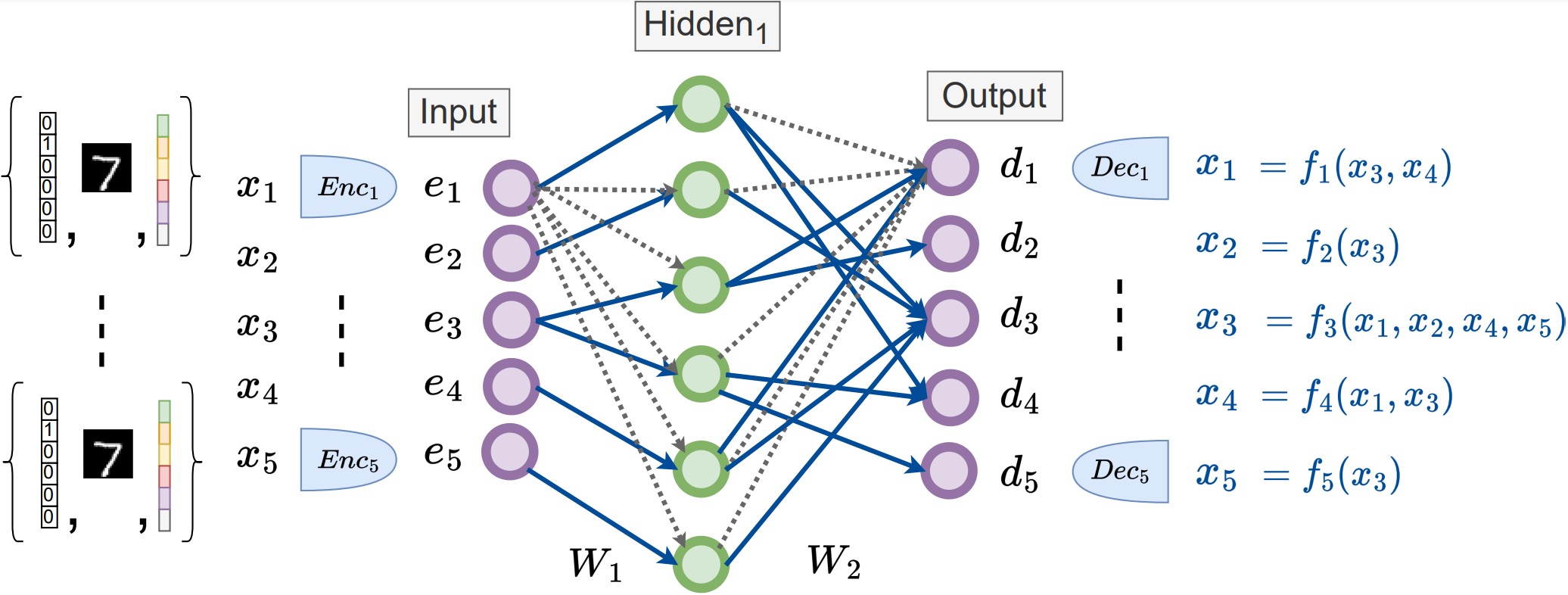}
\caption{\small \textbf{\ngms with projection modules}: The input \textbf{X} can be one-hot (categorical), image or generic embedding (text, audio, speech, etc.). Projection modules (encoder + decoder) are used as a wrapper around the neural view of \ngmsns.
%are shown. 
The architecture choice of the projection modules depends on the input data type and users' design choices. 
% Note that the output of the encoder can be more than 1 unit ($e_1$ can be a hypernode). 
The remaining details are similar to Fig.~\ref{fig:neural-view}.}
\label{fig:neural-view-generic}
\vspace{-4mm}
\end{figure*}

\textit{(I) Projection modules.} 
We add a \textit{Projection} module consisting of an encoder and decoder that act as a wrapper around the neural view of the \ngmns, refer to Fig.~\ref{fig:neural-view-generic}. Without loss of generality we consider that each of the $D$ inputs is an embedding in $x_i\in\mathbb{R}^I$. We convert all the input $x_i$ nodes in the \ngm architecture to hypernodes, where each hypernode contains the embedding vector. Consider a hypernode that contains an embedding vector of size $E$ and if an edge is connected to the hypernode, then that edge is connected to all the $E$ units of the embedding vector. For each of the input hypernodes, we define a corresponding encoder embedding $e_i\gets\text{enc}_i(x_i),\forall e_i\in\mathbb{R}^E$, which can be designed specifically for that particular input embedding. We apply the encoder modules to all the $x_i$ hypernodes and obtain the $e_i$ hypernodes. Same procedure is followed at the decoder end, where $x_i\gets\text{dec}_i(d_i), \forall d_i\in\mathbb{R}^O$. 
%We do this step primarily to reduce the input dimensions. 
Now, the \ngm optimization reduces to learning the connectivity pattern using the path norms between hypernodes $e_i$'s and $d_i$'s. A slight adjustment to the graph-adjacency matrix $S^c\in\{0, 1\}^{DE\times DO}$ will account for the hypernodes. The optimization becomes
%\begin{equation}
\vspace{-2mm}
\begin{multline}
\label{eqn:optimization-function-generic}
    \argmin_{\mathcal{W,B}, \operatorname{proj}} \sum_{k=1}^{M} \norm{X^k - f_{\mathcal{W,B}}(\operatorname{proj}(X^k))}^2_2 
    + \lambda \log{\left(\norm{\left(\Pi_{i=1}^L |W_i|\right) * S^c}_1\right)}
\end{multline}
The projection modules can be jointly learned in the optimization, as shown in Eq.~\ref{eqn:optimization-function-generic}, or alternatively, one can add fine-tuning layers to their pretrained versions as desired.

\textit{(II) Extending soft-graph constraint.} We can view the connections between the $D$ hypernodes of the input embedding $x_i\in\mathbb{R}^I$ to the corresponding input of the encoder layer $e_i\in\mathbb{R}^E$ as a graph. We represent the set of input layer to the encoder layer connections by $S_{\text{enc}}\in\{0, 1\}^{DI\times DE}$, where there is a $S_{\text{enc}}[x_i, e_j]=1$ if the $(x_i, e_j)$ hypernodes are connected. If we initialize a fully connected neural network (or MLP) between the input layer and the encoder layer, we can utilize the soft-graph penalty to map the paths of the input units to the encoder units in order to satisfy the graph structure defined by $S_{\text{enc}}$. Similarly for the decoder we obtain $S_{\text{dec}}$. We get the following Lagrangian based optimization by extending soft-graph constraints to the connection patterns of the encoder and decoder networks.
% \vspace{-8mm}
\begin{equation}
\begin{split}
\label{eqn:optimization-function-generic-soft}
    &\argmin_{\mathcal{W,B},\mathcal{W}^e, \mathcal{W}^d} \sum_{k=1}^{M} \norm{X^k - f_{\mathcal{W,B},\mathcal{W}^e,\mathcal{W}^d}(X^k)}^2_2 
    + \lambda \log{\left(\norm{\left(\Pi_{i=1}^L |W_i|\right) * S^c}_1\right)} \\ 
    &\qquad+ \lambda_e \log{\left(\norm{\left(\Pi_{i=1}^{L_e} |W^e_i|\right) * S_{enc}^c}_1\right)} 
    + \lambda_d \log{\left(\norm{\left(\Pi_{i=1}^{L_d} |W^d_i|\right) * S_{dec}^c}_1\right)}
\end{split}
\end{equation}
% \vspace{-2mm}
where $f_{\mathcal{W,B},\mathcal{W}^e,\mathcal{W}^d}(\cdot)$ represents the entire end-to-end MLP including the encoder and decoder mappings. The Lagrangian constants $\lambda, \lambda_e, \lambda_d$ are initialized in the same manner as explained in Sec.~\ref{sec:ngm-learning}.  We recommend this approach as training is efficient, highly scalable and can handle large embedding sizes by leveraging batch processing.

\subsection{Inference}\label{sec:ngm-inference}

Inference is the process of using the graphical model to answer queries. Calculation of conditional distributions and maximum a-posteriori (MAP) values are key operations for inference. 
\ngm marginals can be obatined using the frequentist approach from the input data. % or using the feature priors introduced in the previous section.
% As an example
\begin{wrapfigure}[32]{R}{0.63\textwidth}
%\vspace{-2mm}
\begin{algorithm}[H]
  \DontPrintSemicolon
  %\SetKwFunction{Grad}{Grad}
  \SetKwProg{Fn}{Function}{:}{}
  \SetKwFor{uWhile}{while}{do}{}
  \SetKwRepeat{Do}{do}{while}
  \SetKwFor{ForPar}{For all}{do in parallel}{}
  \SetKwFunction{messagePassing}{message-passing}
  \SetKwFunction{gradientbased}{gradient-based}
  \SetKwFunction{inference}{ngm-inference}
    \Fn{\gradientbased{$f_\mathcal{W}, X_I$}}{
        $\{K, U\} \gets I$, split the variables\;%\\
        $K\gets$ fixed indices (known)\;%\\
        $U\gets$ learnable indices (unknown)\;%\\
        $f_\mathcal{W}\gets$ freeze weights\;%\\
      \Do{$\mathcal{L}_\text{In} > \epsilon$}{
            $X_O$ = $f_\mathcal{W}(X_I)$\;%\\
            $\mathcal{L}_\text{In}$ = $\norm{X_O[K] - X_I[K]}_2^2$\;%~ $k$=known indices\;
            $X_I[U] \gets $ update learnable tensors by gradient descent on $\mathcal{L}_\text{In}$\;
        }
     \KwRet $X_I[U]$
    }\vspace{1mm}
    \Fn{\messagePassing{$f_\mathcal{W}, X_I^0$}}{
        \{$K, U\} \gets I$, split the variables  \;%\\
        $t=0$\;%\\
      \uWhile{$\norm{X_I^t - X_I^{t-1}}_2^2>\epsilon$}{
            $\{X_I^{t}[U]; X_I[K]\}$ = $f_\mathcal{W} (\{X_I^{t-1}[U]; X_I[K]\})$\;%\\
            $t = t+1$\;
        }
     \KwRet $X_I^t[U]$
    }\vspace{1mm}
    \Fn{\inference{$f_\mathcal{W}, X_I^0$}}{
        Input: $f_\mathcal{W}$ trained \ngm model\;%\\
        ~$X_I^0\in\mathbb{R}^{D\times 1}$ (mean values for unknown)\;%\\
        $X_I \gets $\messagePassing($f_\mathcal{W}, X_I^0$)\;%\\
        $~~~~~~ \cdots$ or $\cdots$ \;%\\
        $X_I \gets $\gradientbased($f_\mathcal{W}, X_I^0$)\;%\\
        \KwRet $X_I$
    }%\vspace{1mm}
\caption{\ngmsns: Inference algorithm}\label{algo:ngm-inference}
\end{algorithm}
\end{wrapfigure}

We consider two iterative procedures to answer conditional distribution queries over \ngms described in Alg.~\ref{algo:ngm-inference}. 
We split the input variables $\{K, U\} \gets I$ into two parts, $K$ denotes the variables with known (observed) values and $U$ denotes the unknown (target) variables. The inference task is to predict the maximum a posteriori (MAP) values of the unknown nodes based on the trained \ngm model distributions. In the fist approach, we use the popular message passing algorithms that keeps the observed values of the features fixed and iteratively updates the values of the unknowns until convergence. We developed an alternative algorithm which is efficient and is our recommended approach to perform inference in \ngmsns.

\textbf{Gradient based approach to computing MAP values}:
The weights of the trained \ngm model are frozen once trained. 
% repetition
%The input data is divided into fixed $K$ (observed) and learnable $U$ (target) variable indices. 
We define the regression loss over the known attribute values as we want to make sure that the prediction matches values for the observed features. Using this loss we update the learnable input tensors alternating between forward and backward passes until convergence to obtain the values of the target features. Note that the backward pass shares its reliance on gradient with backpropagation, but in our procedure, only node values are updated, the weights remain frozen.
Since the \ngm model is trained to match the output $O$ to the input $I$, 
we can view this procedure of iteratively updating the unknown features so that the input and output matches. 
Based on the convergence loss value reached after the optimization, one can assess the confidence in the inference. Furthermore, plotting the individual feature dependency functions also helps in gaining insights about predicted values. 

\textbf{Obtaining conditional probability distributions.} It is often desirable to get the full conditional probability density function rather than just a point value for any inference query. In case of categorical variables, this is readily obtained as we output a distribution over all the categories (using one-hot encoding). In practice, given a distribution over different categories obtained during inference, we clip the individual values between $[\epsilon, 1]$\footnote{$\epsilon$ is an arbitrarily small value used to avoid setting any probability value to 0.} and then divide by the total sum to get the final distribution. For numerical features, we consider a binned input and corresponding real valued output. The input node corresponding to the numerical feature is split into $m$ nodes, each corresponding to one bin. This is similar to converting the feature to a multi-valued categorical variable.  The output node for the feature remains unsplit.  We link each bin-node to retain the paths learned in training for the original feature.  With this slight modification, the regression term of the loss function Eq.~\ref{eqn:optimization-function-generic} 
%will take binned input and output a real value for the real valued features
becomes $\sum_{k=1}^{M} \norm{X_{\mathcal{O}\text{-real}}^k - f_{\mathcal{W}}(\operatorname{proj}(X_{\mathcal{I}\text{-binned}}^k))}^2_2$.

% \textbf{Sampling}: 
% An \ngm model can also provide efficient sampling utilizing the inference mechanism above. Details are provided in Appendix~\ref{sec:ngm-sampling}.

\section{Sampling}\label{sec:ngm-sampling}

\begin{wrapfigure}[20]{R}{0.63\textwidth}
% \begin{figure}
\vspace{-8mm}
%    \centering
\begin{algorithm}[H]
\caption{\ngms: Sampling algorithm}
\label{algo:ngm-sampling}
  \DontPrintSemicolon
  %\SetKwFunction{Grad}{Grad}
  \SetKwProg{Fn}{Function}{:}{}
%   \SetKwFor{uWhile}{while}{do}{}
  \SetKwFor{uFor}{For}{do}{}
  \SetKwRepeat{Do}{do}{while}
  \SetKwFor{ForPar}{For all}{do in parallel}{}
  \SetKwFunction{getSample}{get-sample}
  \SetKwFunction{sampling}{ngm-sampling}
    \Fn{\getSample{$f_\mathcal{W}, \mathcal{D}_s$}}{ 
        $D$ = len($\mathcal{D}_s$)\;% \\
        $X\in\mathbb{R}^{D\times 1}$ (init learnable tensor)\;% \\
        % Sample $1^{st}$ feature value from empirical marginal distribution $x_1\sim \mathcal{U}(P(x_1|\textbf{X}))$ \;
        Sample $1^{st}$ feature value from empirical marginal distribution $x_1\sim \mathcal{U}(P(x_1))$ \; %\\
        \uFor{$i = 2,\cdots, D$}{
            %X[i] = X[i] + $\epsilon_i$ (add random noise)\;$ \\
            $K\gets 1:i-1$ (fixed tensor indices)\; %\\
            $U\gets i:D$ (learnable tensor indices)\;% \\
            % $X\gets \{X_k, X_u\}$\;% \\
            $P(x_i|X[K])\gets\operatorname{\ngmns-inference}(f_\mathcal{W}, \{X[K], X[U]\})$\; %\\
            $X[i]\sim\mathcal{U}\left(P(x_i|X[K])\right)$ %\\
            }
     \KwRet $X$
    }
    \vspace{1mm}
    \Fn{\sampling{$f_\mathcal{W}$, \textbf{G}}}{
        Input: $f_\mathcal{W}$ trained \ngm model\;% \\
        Randomly choose $x_i$'th starting feature\;% \\
        $\mathcal{D}_s$=BFS(\textbf{G},$x_i$) [undirected]\;% \\
        \quad $\cdots$ queue the features $\cdots$ \;% \\
        $\mathcal{D}_s$=topological-sort(\textbf{G}) [DAGs]\;% \\
        $X \gets $\getSample($f_\mathcal{W}, \mathcal{D}_s$)\;% \\
        \KwRet $X$ 
    }%\vspace{1mm}
\end{algorithm}%\vspace{-5mm}
% \end{figure}
\end{wrapfigure}

To sample from the \ngm 
we propose a procedure akin to forward sampling in Bayesian networks described in Alg.~\ref{algo:ngm-sampling}. 
We based our sampling procedure to follow 
%$X_i=f_{nn}(\text{nbrs}(X_i)) +\epsilon$, where $\epsilon\sim\mathcal{P}$ is random noise. 
$X_i\sim \mathcal{U}(f_{nn}(\text{nbrs}(X_i)))$.
Note that nbrs$(X_i)$ will be MB$(X_i)$ for DAGs. 
To get each sample, we start by choosing a feature at random.
% either at random or based on the structure on the graph :-)
% and based on the structure of the graph. 
To get the order in which the features will be sampled, we do a breadth-first-search (topological sort in DAGs) and arrange the nodes in $\mathcal{D}_s$. In this way, the immediate neighbors are chosen first and then the sampling spreads over the graph away from the starting feature. 
%As we go through the ordered features in the sampling procedure, we 
%sample the value of each feature from the conditional distribution based on previously assigned values
We start by sampling the value of the first feature from the empirical marginal distribution.  We keep it fixed for the subsequent iterations (feature is now observed). 
We then call the inference algorithm conditioned on this fixed feature value to get the 
%values of
distributions over
the unknown features. We sample the value of each subsequent feature in the ordering from the conditional distribution based on previously assigned values. This process is repeated till we get a sample value of all the features. 
The conditional updates are defined as $p\left(X_i^{k}, X_{i+1}^{k}, \cdots, X_D^{k}|X_1^{k}, \cdots, X_{i-1}^{k} \right)$. We keep on fixing the values of features
%(with a small added noise) 
and run inference on the remaining features until we have obtained the values for all the features and thus get a new sample. 

The inference algorithm of the \ngm facilitates conditional inference on multiple unknown features over multiple observed features. Furthermore, all the \ngm algorithms above can be executed in batch mode. We leverage these capabilities of the inference algorithm for faster sampling from \texttt{NGMs}. We can sample from a conditional distribution by pre-setting the values of known variables and update conditional distributions with both pre-set and already instantiated values as given.

\section{Experiments}\label{sec:exp}
We evaluate \ngms on synthetic and real data. 
%Appendix~\ref{apx:design-strategies} contains some best practices that we developed while working with \ngmsns. 
In this section, we cover experiments on infant mortality data.  Additional details and graphs for this domain are included in  Appendix~\ref{apx:infant_mortality_extra}.  In Appendix~\ref{apx:GGM} we show experiments on Gaussian Graphical models and in Appendix~\ref{apx:lung-cancer} on lung cancer data. 
We discuss design strategies and optimization details for \ngms in Appendix~\ref{apx:design-strategies}.

\textbf{Infant mortality analysis}: 
%We created an \ngm to model infant mortality data. 
The dataset is based on CDC Birth Cohort Linked Birth – Infant Death Data Files \cite{CDC:InfantLinkedDatasets}.  It describes pregnancy and birth variables for all live births in the U.S. together with an indication of an infant's death before the first birthday.  We used the data for 2015 (latest available), which includes information about 3,988,733 live births in the US during 2015 calendar year.  

\textbf{Recovered graphs}: 
We recovered the graph strucure of the dataset using \uglad \cite{shrivastava2022uglad} and using Bayesian network package \texttt{bnlearn}~\cite{bnlearn} with Tabu search and AIC score.  The graphs are shown in Fig.~\ref{fig:im2015-ci-graph} and \ref{fig:im2015-bn-graph} in the Appendix. Since \texttt{bnlearn} does not support networks containing both continuous and discrete variables, all variables were converted to categorical for \texttt{bnlearn} structure learning and inference. In contrast, \uglad and \ngms are both equipped to work with mixed types of variables and were trained on the dataset prior to conversion.

%\begin{wraptable}[4]{R}{1\textwidth}
\begin{table*}[]
\centering
% \vspace{-2mm}
\caption{\small Comparison of predictive accuracy for gestational age and birthweight.} 
\label{tab:inference_results1}
\resizebox{0.8\textwidth}{!}{
\begin{tabular}{|c|c|c|c|c|}
\hline
Methods & \multicolumn{2}{c|}{Gestational age} & \multicolumn{2}{c|}{Birthweight}  \\ 
& \multicolumn{2}{c|}{(ordinal, weeks)} & \multicolumn{2}{c|}{(continuous, grams)}  \\ \hline
& MAE & RMSE & MAE & RMSE \\ \hline
Logistic Regression & $1.512\pm0.005$ & $3.295\pm0.043$ & N/A & N/A  \\ \hline
Bayesian network & {\boldmath $1.040 \pm0.003$} & $2.656\pm0.027$ & N/A & N/A \\ \hline
EBM & $1.313 \pm0.002$ & $2.376\pm0.021$ & { \boldmath $345.21\pm1.47$ } & {\boldmath $451.59\pm2.38$}  \\ \hline
\ngm w/full graph & $1.560 \pm0.067$ & $2.681\pm0.047$ & $394.90\pm11.25$ & $517.24\pm11.51$  \\ \hline
\ngm w/BN graph & $1.364 \pm0.025$ & $2.452\pm0.026$ & $370.20\pm1.44$ & $484.82\pm1.88$  \\ \hline
\ngm w/\uglad graph & $1.295 \pm0.010$ & {\boldmath $2.370\pm0.025$ } & $371.27\pm1.78$ & $485.39\pm1.86$  \\ \hline
\end{tabular}}
\label{tab:inference_results}
\vspace{-4.5mm}
%\end{wraptable}
\end{table*}

%\begin{wraptable}[4]{R}{1\textwidth}
\begin{table*}[]
\centering
\vspace{-2mm}
\caption{\small Comparison of predictive accuracy for 1-year survival and cause of death.  Note: recall set to zero when there are no labels of a given class, and precision set to zero when there are no predictions of a given class.} 
\resizebox{\textwidth}{!}{
\begin{tabular}{|c|c|c|c|c|c|c|}
\hline
Methods & \multicolumn{2}{c|}{Survival} & \multicolumn{4}{c|}{Cause of death} \\ 
& \multicolumn{2}{c|}{(binary)} & \multicolumn{4}{c|}{(multivalued, majority class frequency $0.9948$)} \\ \hline
& \multicolumn{2}{c|}{} & \multicolumn{2}{c|}{micro-averaged}  & \multicolumn{2}{c|}{macro-averaged}  \\ 
& AUC & AUPR & Precision & Recall & Precision & Recall\\ \hline
Logistic Regression & $0.633\pm0.004$ & $0.182\pm0.008$ & $0.995\pm7.102\text{e-}05$ & $0.995\pm7.102\text{e-}05$ & $0.136\pm0.011$ & $0.130\pm0.002$ \\ \hline
Bayesian network & $0.655\pm0.004$ & $0.252\pm0.007$ & $0.995\pm7.370\text{e-}05$  & $0.995\pm7.370\text{e-}05$ &  $0.191\pm0.008$ & $0.158\pm0.002$\\ \hline
EBM & $0.680\pm0.003$ & {\boldmath $0.299\pm0.007$ } & {\boldmath $0.995\pm5.371\text{e-}05$} & {\boldmath $0.995\pm5.371\text{e-}05$} &  $0.228\pm0.014$ & $0.166\pm0.002$ \\ \hline
\ngm w/full graph & $0.721\pm0.024$ & $0.197\pm0.014$ & $0.994\pm1.400\text{e-}05$ & $0.994\pm1.400\text{e-}05$ & {\boldmath $0.497\pm7.011\text{e-}06$} & {\boldmath $0.500\pm1.000\text{e-}06$ }\\ \hline
\ngm w/BN graph & {\boldmath $0.752\pm0.012$ } & $0.295\pm0.010$ & $0.995\pm4.416\text{e-}05$ & $0.995\pm4.416\text{e-}05$ & {\boldmath $0.497\pm2.208\text{e-}05$} & {\boldmath $0.500\pm1.000\text{e-}06$ }\\ \hline
\ngm w/\uglad graph & $0.726\pm0.020$ & $0.269\pm0.018$ & $0.995\pm9.735\text{e-}05$ & $0.995\pm9.735\text{e-}05$ & $0.497\pm4.868\text{e-}05$ & {\boldmath $0.500\pm1.000\text{e-}06$} \\ \hline
\end{tabular}}
\label{tab:inference_results2}
\vspace{-5mm}
%\end{wraptable}
\end{table*}

\textbf{\ngms trained on infant mortality dataset}: 
%\textbf{\ngm generic architecture}: 
Since we have mixed input data types, real and categorical data, we utilize the \texttt{NGM}-generic architecture, refer to Fig.~\ref{fig:neural-view-generic}. We used a 2-layer neural view with $H=1000$. The categorical input was converted to its one-hot vector representation and %clubbed together with 
added to
the real features which gave us roughly $\sim 500$ features as input, see Appendix~\ref{apx:infant_mortality_extra}. 
\ngm was trained on the 4 million data points with $\sim500$ features using 128 CPUs within 2 hours.  
% The \texttt{NGM}-generic parameters were learned by optimizing Eq.~\ref{eqn:optimization-function-generic-soft}.% using the adam optimizer. 

\textbf{Inference accuracy comparison}: Infant mortality dataset is particularly challenging, since cases of infant death during the first year of life are (thankfully) rare.  Thus, any queries concerning such low probability events are hard to estimate with accuracy.
To evaluate inference accuracy of \ngmsns, we compared prediction for four variables of various types:  gestational age (ordinal, expressed in weeks), birthweight (continuous, specified in grams), survival till 1st birthday (binary) and cause of death ("alive", 10 most common causes of death with less common grouped in category "other" with "alive" indicated for 99.48\% of infants).  For each case, the dataset was split randomly into training and test sets (80/20) 20 times, each time a model was trained on the training set and accuracy metrics evaluated on the test set.  We compared the performance of logistic regression, Bayesian networks, Explainable Boosting Machines (EBM)~\cite{caruana2015intelligible,lou2013accurate} and \ngmsns.  In case of \ngmsns, we trained two models: one using the Bayesian network graph and one using the uGLAD graph.

% Our experiment's results are presented in 
Tables~\ref{tab:inference_results1} and~\ref{tab:inference_results2} demonstrate that \ngm models are significantly more accurate than logistic regression, more accurate than Bayesian Networks and on par with EBM models for categorical and ordinal variables. They particularly shine in predicting very low probability categories for multi-valued variable cause of death, where most models (both PGMs and classification models) typically struggle. Note that we need to train a separate EBM model for each outcome variable evaluated, while all variables can be predicted within one trained \ngm model. 
Interestingly, the two \ngm models show similar accuracy results despite the differences in the two dependency structures used in training.

Our experiments on infant mortality dataset demonstrate usefulness of \ngms to model complex mixed-input real-world domains.  We are currently running more experiments designed to capture more information on \ngmsns' sensitivity to input graph recovery algorithm and its impact on inference accuracy.

\section{Conclusions}\label{sec:conclusions}
This work attempts to improve the usefulness of Probabilistic Graphical Models by extending the range of input data types and distribution forms such models can handle. Neural Graphical Models provide a compact representation for a wide range of complex distributions and support efficient learning, inference and sampling. The experiments are carefully designed to systematically explore the various capabilities of \ngmsns. Though \ngms can leverage GPUs and distributed computing hardware, we do forsee some challenges in terms of scaling in number of features and performance on very high-dimensional data. Using \ngms for images and text based applications will be interesting to explore. We believe that \ngms is an interesting amalgam of the deep learning architectures' expressivity and Probabilistic Graphical Models' representation capabilities. 

\bibliographystyle{ECSQARUfiles/splncs04}
\bibliography{citations,bibfile}

\clearpage
\appendix

\section{Design strategies and best practices for \ngms}\label{apx:design-strategies}
We share some of the design strategies and best practices that we developed while working with \ngms in this section. This is to provide insights to the readers on our approach and help them narrow down the architecture choices of \ngms for applying to their data. We hope that sharing our thought process and findings here will foster more transparency, adoption, and help identify potential improvements to facilitate the advancement of research in this direction.
\begin{itemize}[leftmargin=*,nolistsep]
    \item \textit{Choices for the structure loss function.} We narrowed down the loss function choice to Hadamard loss $\norm{\left(\Pi_i |W_i|\right) * S^c}$ vs square loss $\norm{\left(\Pi_i |W_i|\right) - S}^2$. We also experimented with various choices of Lagrangian penalties for the structure loss. We found that $\ell_2$ worked better in most cases. Our conclusion was to use Hadamard loss with either $\ell_1$ vs $\ell_2$ penalty.
    \item \textit{Strategies for $\lambda$ initialization.} (I) Keep it fixed to balance between the initial regression loss and structure loss. We utilize the loss balance technique mentioned in~\cite{rajbhandari2019antman}. (II) Use the proximal initialization technique combined with increasing $\lambda$ value as described in Alg.~\ref{algo:ngms-learning}. Both techniques seem to work well, although (I) is simpler to implement and gives equivalent results.
    \item \textit{Selecting width and depth of the neural view.} We start with hidden layer size $H=2\times|I|$, that is, twice the input dimension. Then based on the regression and structure loss values, we decide whether to go deeper or have a larger number of units. In our experience, increasing the number of layers helps in reducing the regression loss while increasing the hidden layer dimensions works well to optimize for the structure loss. 
    \item \textit{Choice of non-linearity.} For the MLP in the neural view, we played around with multiple options for non-linearities. We ended up using ReLU, although $\tanh$ gave similar results.
    \item \textit{Handling imbalanced data.} \ngms can also be adapted to utilize the existing imbalanced data handling techniques~\cite{chawla2002smote,shrivastava2015classification,bhattacharya2017icu,bhattacharya2019methods,shrivastava2021system} which improved results in our experience.  Note excellent results for a multi-valued categorical variable where the majority class probability exceeds 99\%  (Section~\ref{sec:exp}).
    \item \textit{Calculate upper bound on regression loss.} Try fitting \ngm by assuming fully connected graph to give the most flexibility to regression. This way we get an upper bound on the best optimization results on just the regression loss. This helps to select the depth and dimensions of MLPs required when the sparser structure is imposed. 
    \item \textit{Convergence of loss function.} In our quest to figure out a way to always get good convergence on both the losses (regression \& structure), we tried out various approaches. (I) Jointly optimize both the loss functions with a weight balancing term $\lambda$, Eq.~\ref{eqn:optimization-function}. (II) We tested out Alternating Method of Multipliers (ADMM) based optimization that alternately optimizes for the structure loss and regression loss. (III) We also ran a proximal gradient descent approach which is sometimes suitable for loss with $\ell_1$ regularization terms. Choice (I) turned out to be effective with reasonable $\lambda$ values. The recommended range of $\lambda$ is [1e-2, 1e2]. 
\end{itemize}

In the current state, it can be tedious to optimize \ngms and it requires a fair amount of experimentation. It is a learning experience for us as well and we are always on a lookout to learn new techniques from the research community.

\section{Modeling Gaussian Graphical models}
\label{apx:GGM}
We designed a synthetic experiment to study the capability of \ngms to represent Gaussian graphical models. 
The aim of this experiment is to see (via plots and sampling) how close are the distributions learned by the \ngms  to the GGMs.

\begin{figure}
% \vspace{-3.5mm}
\subfigure{\label{fig:ggm1}\includegraphics[width=0.18\textwidth,height=35mm]{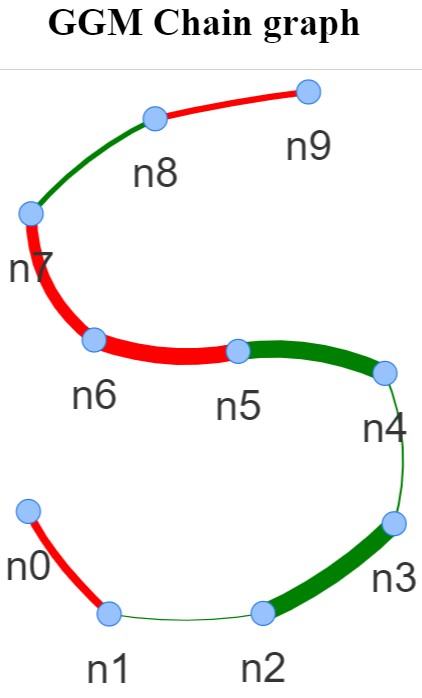}~}
\subfigure{\label{fig:ggm2}\includegraphics[width=0.28\textwidth,height=35mm]{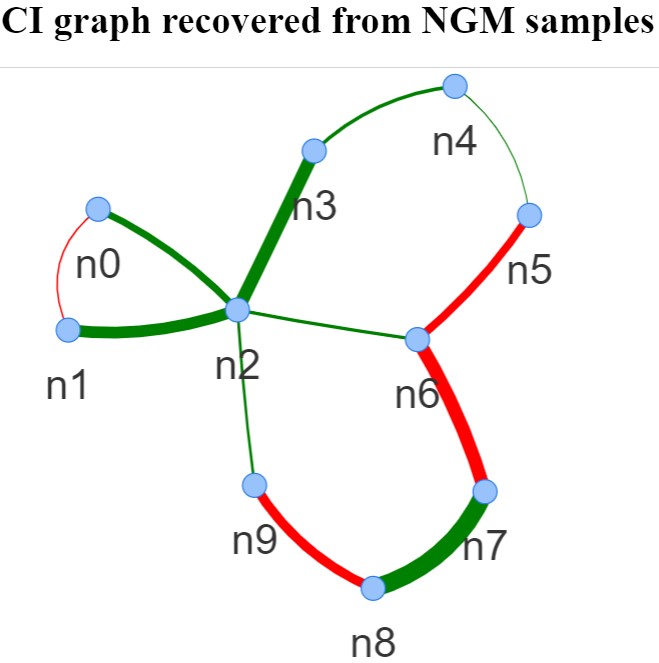}~}
\includegraphics[width=0.45\textwidth,height=35mm]{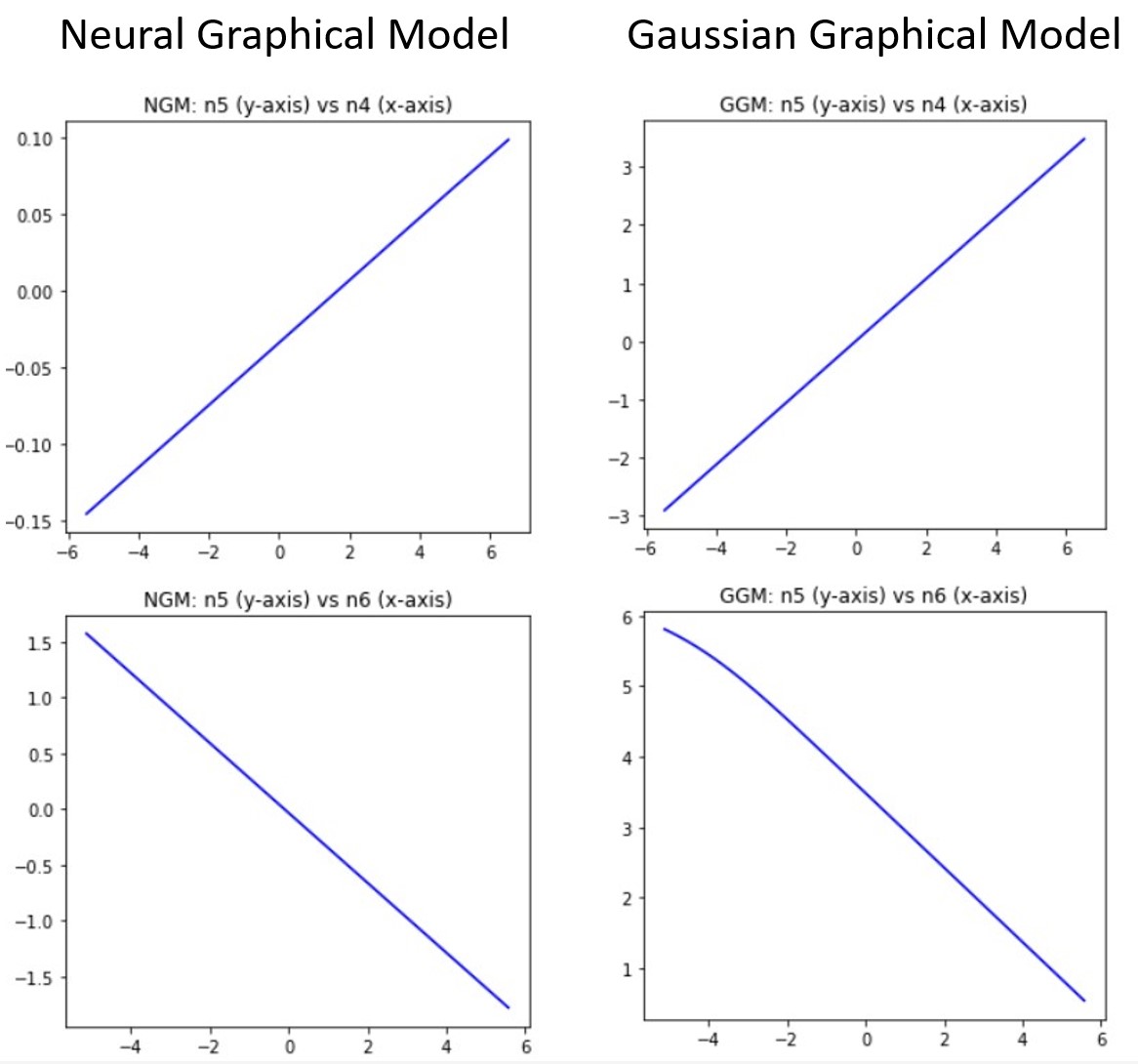}
% \vspace{-6mm}
\caption{\small The graph on the left shows the chain graph  \textbf{G} (partial correlations in green are positive, red are negative, thickness shows the correlations strength) obtained from the initialized partial correlation matrix. Samples $\textbf{X}\in\mathbb{R}^{2000\times 10}$ were drawn from the GGM. \ngm was learned on the input (\textbf{X}, \textbf{G}). The two plots on the right show the dependency functions of \ngm and GGM for a particular node by varying its neighbor's values. 
%As the underlying 
%graph 
%distribution is multivariate Gaussian, 
The positive and negative correlations are reflected in the slope of the curve, as expected analytically. We then sampled from the learned \ngm to obtain data $\textbf{Xs}\in\mathbb{R}^{M_s\times 10}$. 
%We then studied the extent to which the CI graph recovery algorithms  were able to retrieve the underlying graph. 
The middle of the figure shows the graph recovered by running \uglad~\cite{shrivastava2022uglad} on \textbf{Xs}. 
%We can observe that it missed some of the edges but most of the connections along with the correlations signs were retrieved from the \ngm samples.
We can observe that it recovered all the original edges with correct correlation signs. There are three spurious edges not present in the original graph.}
\label{fig:ggm-expt}
\end{figure}
% \end{wrapfigure}

\begin{table}[h]
%\floatbox[{\capbeside\thisfloatsetup{capbesideposition={left, center},capbesidewidth=8.5cm}}]{table}[0.999\FBwidth]
{\caption{\small The CI graph recovered from \ngm samples is compared with the CI graph defined by the GGMs precision matrix. Area under the ROC curve (AUC) and Area under the precision-recall curve (AUPR) values for 10 runs are reported, refer to Fig.~\ref{fig:ggm-expt}.}\label{tab:ggm-expt}}
\centering
\resizebox{0.35\textwidth}{!}{
\centering
\begin{tabular}{|c|c|c|}
\hline
Samples & AUPR & AUC \\ \hline
1000 & $0.84\pm0.03$ & $0.91\pm 0.002$ \\ \cline{1-1}
2000 & $0.86\pm0.02$ & $0.93\pm 0.001$ \\ \cline{1-1}
4000 & $0.96\pm0.00$ & $0.99\pm 0.003$ \\ \hline
\end{tabular}
}

\end{table}

\subsection{Setup}
\textit{Define the underlying graph.} We defined a chain (or path-graph) containing D nodes as the underlying graph. We chose this graph as it allows for an easier study of dependency functions.

\textit{Fit GGM and get samples.} Based on the underlying graph structure, we defined a precision matrix $\Theta$ and obtained its entries by randomly sampling from $\Theta_{i,j}\sim \mathcal{U}\{(-1, -0.5)\cup(0.5, 1)\}$. We then used this precision matrix as the multivariate Gaussian distribution parameters to obtain the input sample data \textbf{X}. We get the corresponding partial correlation graph \textbf{G} by using the formula, %~\cite{algo-for-CI-graphs} 
%\begin{align}
    $$\Rho_{X_i, X_j}.\textbf{X}_{D\backslash{i,j}} =-\frac{\Theta_{i,j}}{\sqrt{\Theta_{i,i} \Theta_{j,j}}}$$.
%\end{align}

\textit{Fit \ngm and get samples.} We fit a \ngm on the input (\textbf{X}, \textbf{G}). We chose $H=30$ with 2 layers and non-linearity $\tanh$ for the neural view's MLP. Training was done by optimizing Eq.~\ref{eqn:optimization-function} for the input, refer to Fig.~\ref{fig:ggm-expt}. Then, we obtained data samples $\textbf{Xs}$ from the learned \ngmns.

\subsection{Analysis}
\textit{How close are the GGM and \ngm samples?} We recover the graph using the graph recovery algorithm \uglad on the sampled data points from \ngms and compare it with the true CI graph. Table~\ref{tab:ggm-expt} shows the graph recovery results of varying the number of samples from \ngmsns. We observe that increasing the number of samples improves graph recovery, which is expected.

\textit{Were the \ngms able to model the underlying distributions?} The functions plot (on the right) in Fig.~\ref{fig:ggm-expt} plots the resultant regression function for a particular node as learned by \ngmns. This straight line with the slope corresponding to the partial correlation value is what we expect theoretically for the GGM chain graph. 
This is also an indication that the learned \ngms were trained properly and reflect the desired underlying relations. Thus, \ngms are able to represent GGM models.

\begin{figure}%[b]
%\centering 
\begin{center}
\subfigure{\includegraphics[width=0.55\textwidth]{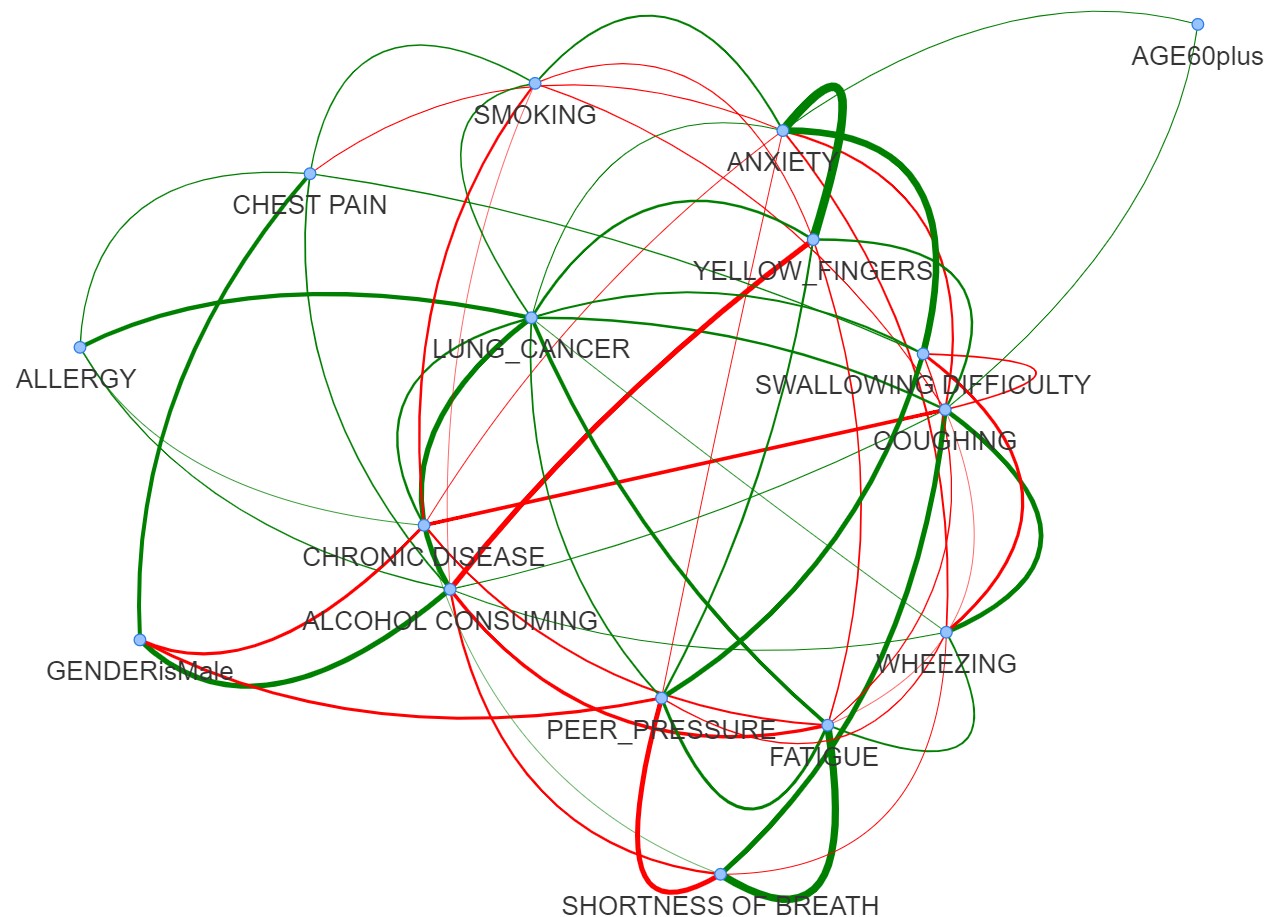}~}\\
\subfigure{\includegraphics[width=0.57\textwidth,height=58mm]{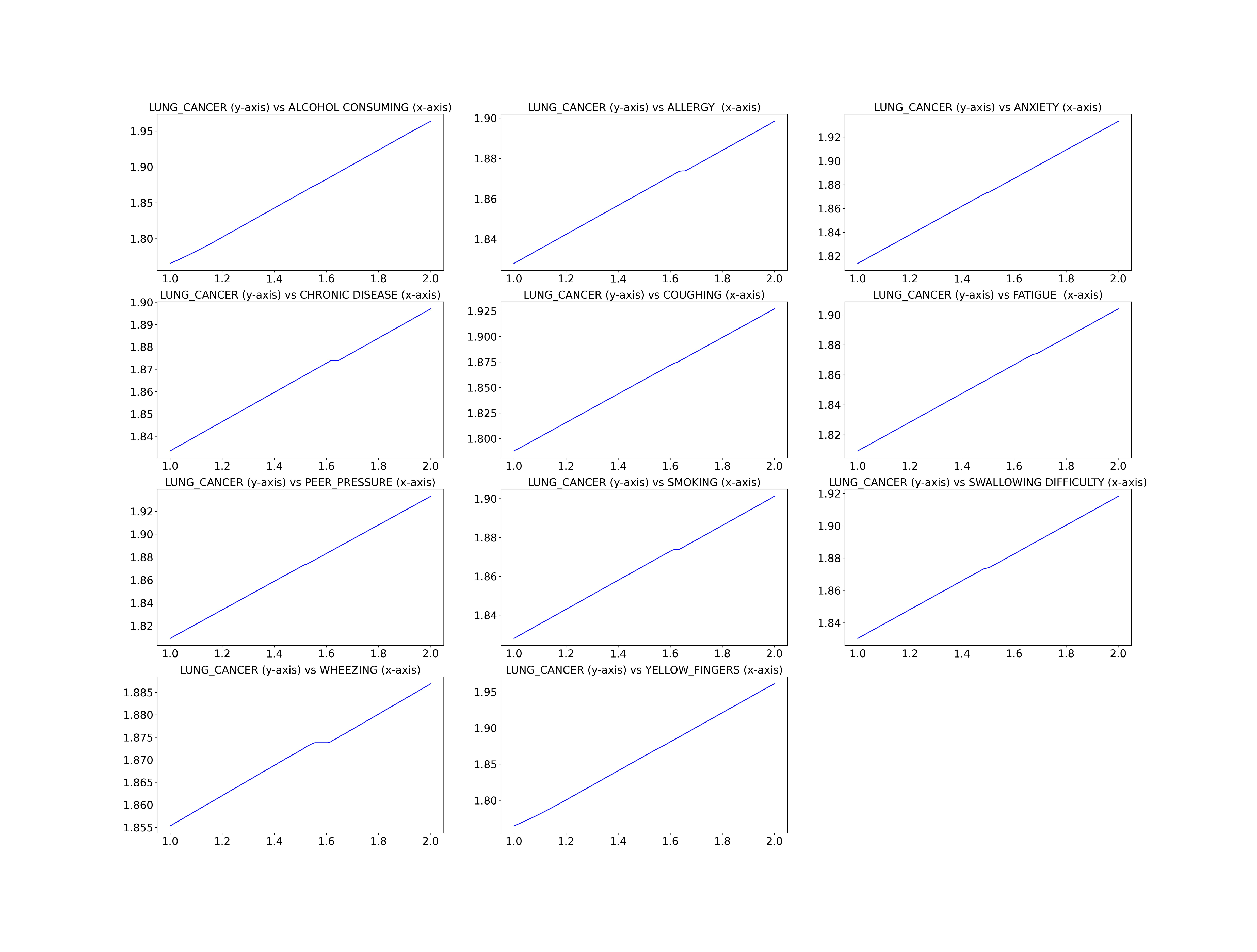}~}
\subfigure{\includegraphics[width=0.38\textwidth,height=58mm]{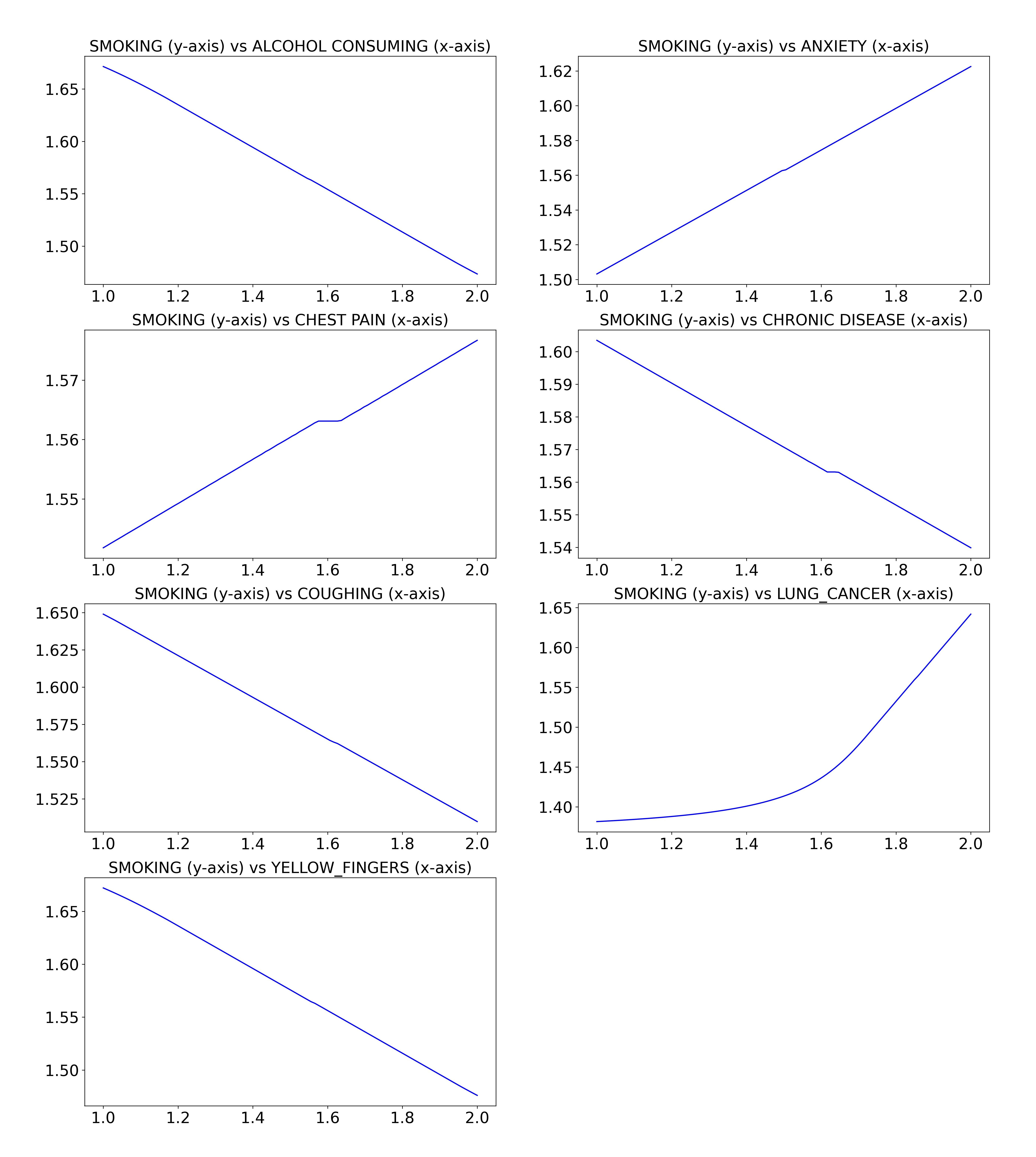}~}
\end{center}
\caption{\small (top) The CI graph recovered by \uglad for the lung cancer data. Plots below show the conditional distribution for the features P(Lung cancer='Yes'| nbrs(Lung cancer)) and P(Smoking| nbrs(Smoking)) based on their neighbors. We used a 2-layer \ngm with hidden size $H=30$ and $\tanh$ non-linearity. \ngms are able to capture non-linear dependencies between the features. Interestingly the \ngms match the relationship trends discovered (positive and negative correlations) by the corresponding CI graph.}
\label{fig:lung-cancer-ci-graph}
\end{figure}

\section{Lung cancer data analysis}
\label{apx:lung-cancer}
We analysed lung cancer data from~\cite{lcData} using \ngmsns. The effectiveness of cancer prediction system helps people to learn their cancer risk with low cost and take appropriate decisions based on their cancer risk status. This data contains 284 patients and for each patient 16 features (Gender, Smoking, Anxiety, Lung cancer present, etc.) are collected. Each entry is a binary entry (YES/NO) or in some cases (AGE), entries are binarized. We used \ngms to study how different features are related and discover their underlying functional dependencies.

%\begin{wraptable}[4]{L}{0.35\textwidth}
%\centering
%\vspace{-3mm}
\begin{table}
\centering
 \caption{\small 5-fold CV results.} 
%\resizebox{\textwidth}{!}{
\begin{tabular}{|c|c|c|}
\hline
Methods & Lung cancer & Smoking \\ \hline
LR & $0.95\pm0.02$ & $0.71\pm0.01$ \\ \hline
\ngm & $0.96\pm0.01$ & $0.79\pm0.02$ \\ \hline
\end{tabular}
\label{tab:reg-lung-cancer}
%\vspace{5mm}
\end{table}
%\end{wraptable}

%Fig.~\ref{fig:lung-cancer-ci-graph} (left) shows the conditional independence graph recovered by running \uglad on the input data. 
The input data along with the CI graph recovered using \uglad were used to learn an \ngm in Fig.~\ref{fig:lung-cancer-ci-graph}. 

In order to gauge the regression quality of \ngmsns, we compare with logistic regression to predict the probability of feature values given the values of the remaining features. Table.~\ref{tab:reg-lung-cancer} shows regression results of logistic regression (LR) and \ngms on 2 different features, \textit{lung cancer} and \textit{smoking}.
%Prediction with a logistic regression model resulted in AUC=$0.951\pm0.02$. The 2-layer \ngm gave AUC=$0.968\pm0.01$. 
The prediction probability for \ngms were calculated by running inference on each test datapoint, eg. P(lung-cancer=yes| $f_i=v_i$ $\forall{i}$ in test data). 
%We note that the data size is not large and is highly unbalanced with regards to positive and negative instances of patients having lung cancer. 
%Similarly, we trained another logistic regression model for the feature `Smoking' with LR giving AUC=$0.711\pm0.01$ and \ngms AUC=$0.79\pm0.02$. 
This experiment primarily demonstrates that a single \ngm model can robustly handle fitting multiple regressions and one can avoid training a separate regression model for each feature while maintaining on par performance. Furthermore, we can obtain the dependency functions that bring in more interpretability for the predicted results, Fig.~\ref{fig:lung-cancer-ci-graph}. Samples generated from this \ngm model can be used for multiple downstream analyses. % and help out in cases where more data points are needed.

\section{~\ngm on Infant Mortality data (details)}
\label{apx:infant_mortality_extra}

\subsection{Representing categorical variables}

Assume that in the input \textbf{X}, we have a column $X_c$ having $\abs{C}$ different categorical entries. One way to handle categorical input is to do one-hot encoding on the column $X_c$ and end up with $\abs{C}$ different columns, $X_c = [X_{c_1}, X_{c_2}, \cdots, X_{c_C}]$. We replace the single categorical column with the corresponding one-hot representation in the original data. The path dependencies matrix \textbf{S} of the MLP will be updated accordingly. Whatever connections were previously connected to the categorical column $X_c$ should be maintained for all the one-hot columns as well. Thus, we connect all the one-hot columns to represent the same path connections as the original categorical column.

\subsection{Additional Infant Mortality results}

\begin{figure}%[b]
\centering 
\includegraphics[width=82mm]{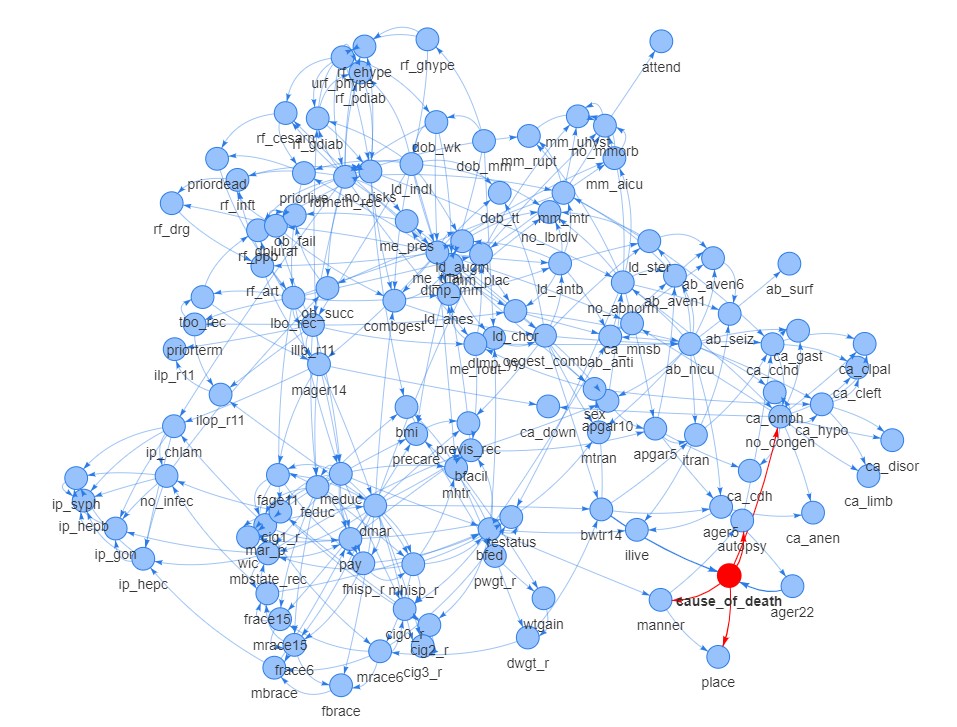}
\caption{\small The Bayesian network graph learned using score-based method for the  Infant Mortality 2015 data. }
\label{fig:im2015-bn-graph}
\vspace{-5mm}
\end{figure}

\begin{figure}%[b]
\centering 
\includegraphics[width=82mm]{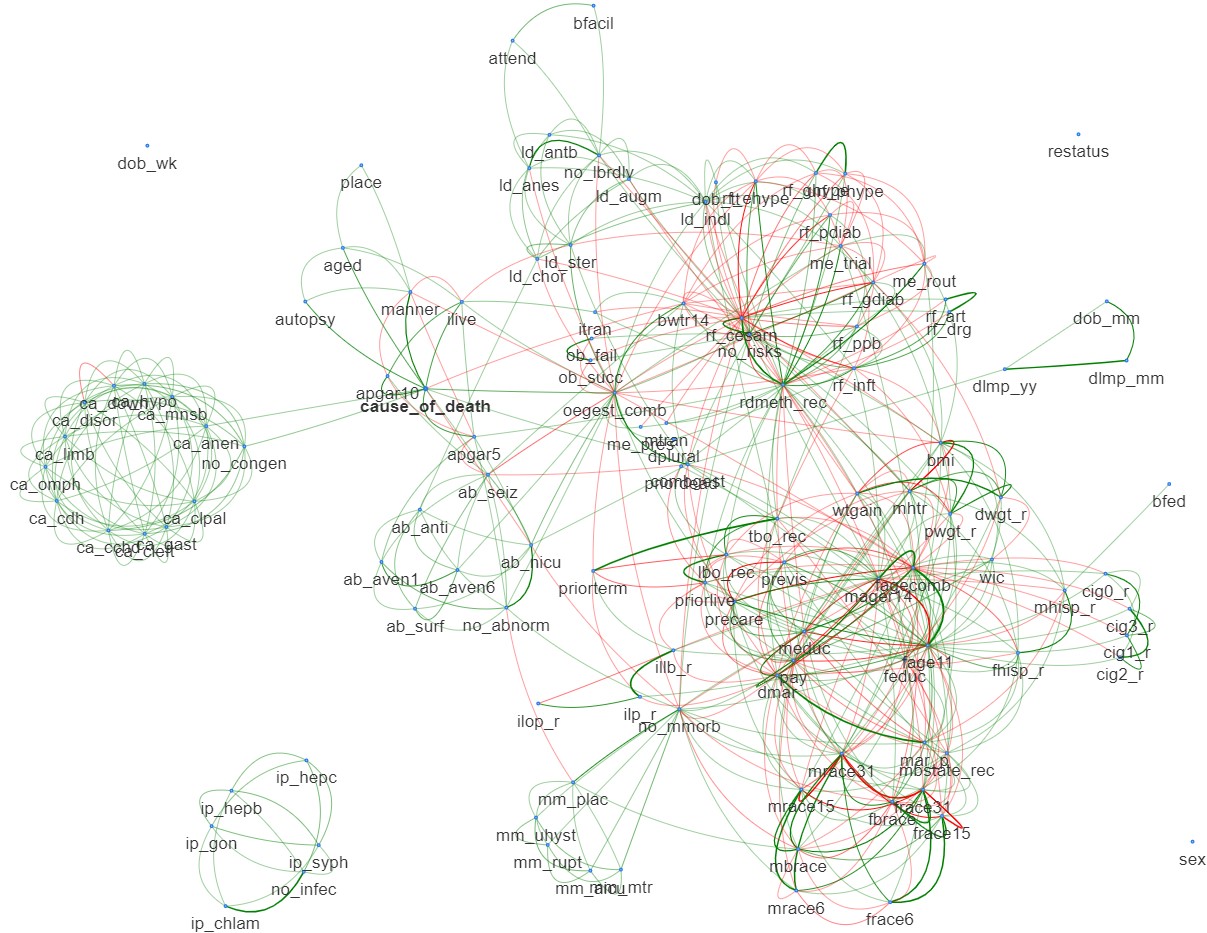}
\caption{\small The CI graph recovered by \uglad for the Infant Mortality 2015 data. 
}
\label{fig:im2015-ci-graph}
\vspace{-5mm}
\end{figure}

\textbf{The dataset and recovered graphs}: 
We recovered the graph strucure of the dataset using \uglad \cite{shrivastava2022uglad} and using Bayesian network package \texttt{bnlearn}~\cite{bnlearn} with Tabu search and AIC score.  The graphs are shown in Fig.~\ref{fig:im2015-ci-graph} and \ref{fig:im2015-bn-graph} respectively. Since \texttt{bnlearn} does not support networks containing both continuous and discrete variables, all variables were converted to categorical for \texttt{bnlearn} structure learning and inference. In contrast, \uglad and \ngms are both equipped to work with mixed types of variables and were trained on the dataset prior to conversion.

Both graphs show similar sets of clusters with high connectivity within each cluster:  
%\vspace{-2mm}
\begin{itemize}[leftmargin=*,nolistsep]
    \item parents' race and ethnicity (\texttt{mrace} \& \texttt{frace}), 
    \item related to mother's bmi, height (\texttt{mhtr}) and weight, both pre-pregnancy (\texttt{pwgt\_r}) and at delivery (\texttt{dwgt\_r}), 
    \item consisting of maternal morbidity variables marked with \texttt{mm} prefix (e.g., unplanned hysterectomy), 
    \item pregnancy related complications such as hypertension and diabetes (variables prefixed with \texttt{rf} and \texttt{urf}), 
%    \item consisting of variables related to parents' STD infections (\texttt{ip} prefix), 
    \item related to delivery complications and interventions (variables prefixed with \texttt{ld}), 
    \item showing interventions after delivery (\texttt{ab} prefix) such as ventilation or neonatal ICU,
    \item describing congenital anomalies diagnosed in the infant at the time of birth (variables prefixed with \texttt{ca}), 
    \item related to infant's death: age, place, autopsy, manner, etc. 
\end{itemize}

Apart from these clusters, there are a few highly connected variables in both graphs: gestational age (\texttt{combgest} and \texttt{oegest}), delivery route (\texttt{rdmeth\_rec}), Apgar score, type of insurance (\texttt{pay}), parents' ages (\texttt{fage} and \texttt{mage} variables), birth order (\texttt{tbo} and \texttt{lbo}), and prenatal care.  

\begin{figure*}%[b]
\centering 
\includegraphics[width=120mm]{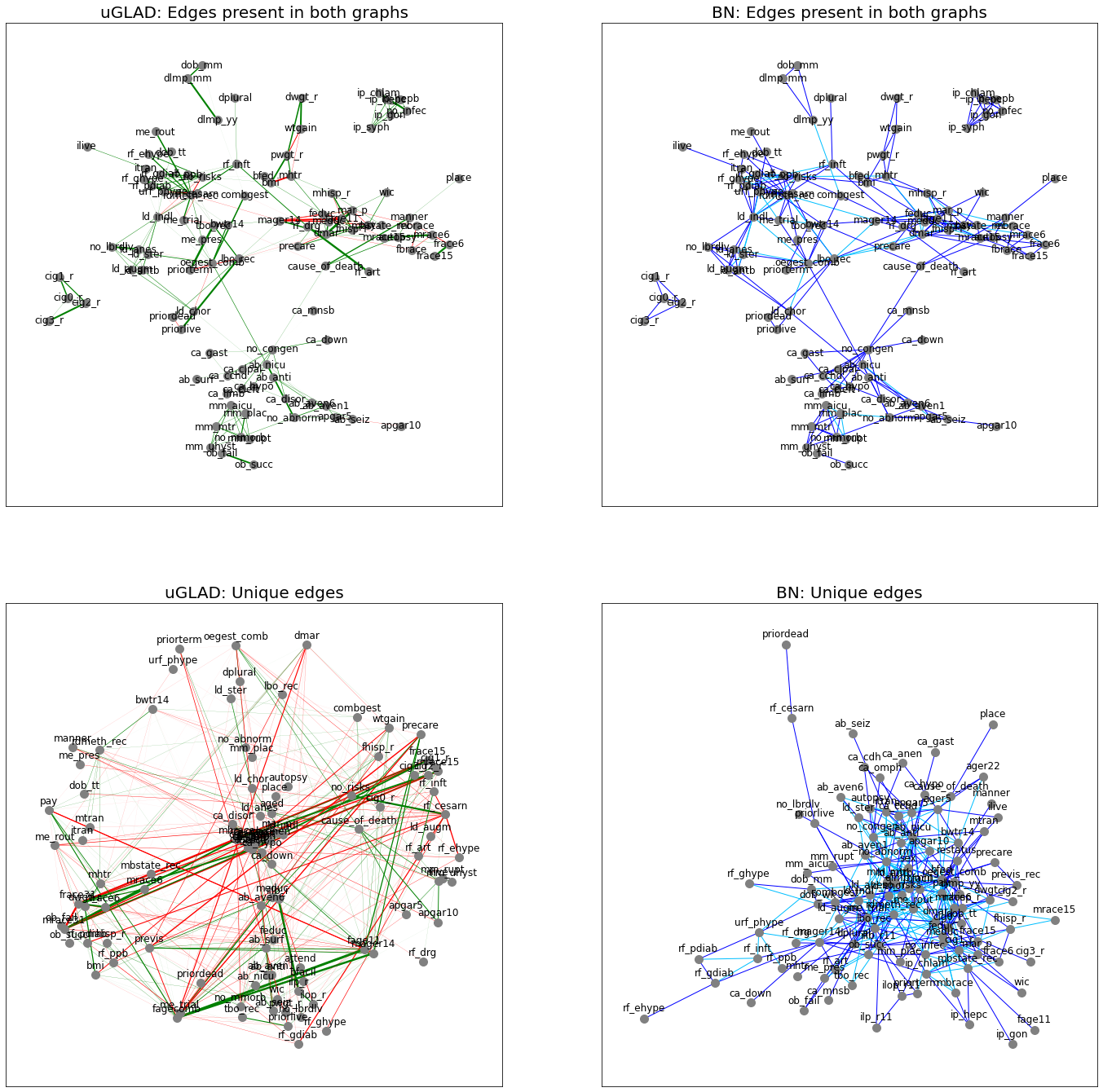}
\caption{\small Comparing the graphs recovered by \uglad and Bayesian Network recovery package~\cite{bnlearn} after moralization (moralized edges are denoted by the skyblue color).}
\label{fig:im2015-compare-graph-bn-moral}
\end{figure*}

With all these similarities, however, the total number of edges varies greatly between the two graphs and the number of edges unique to each graph outnumbers the number of edges the two graphs have in common (see Figure~\ref{fig:im2015-compare-graph-bn-moral}. % in Appendix~\ref{apx:infant_mortality_extra}). % ~\ref{fig:im2015-compare-graph}\&
In particular, most of the negative correlations discovered by \uglad are not present in the BN graph. One reason for the differences lies in the continuous-to-categorical conversion performed prior to Bayesian network structure discovery and training. More importantly, the two graph recovery algorithms are very different in both algorithmic approach and objective function.  

\begin{figure*}%[b]
\centering 
\includegraphics[width=120mm, height=130mm]{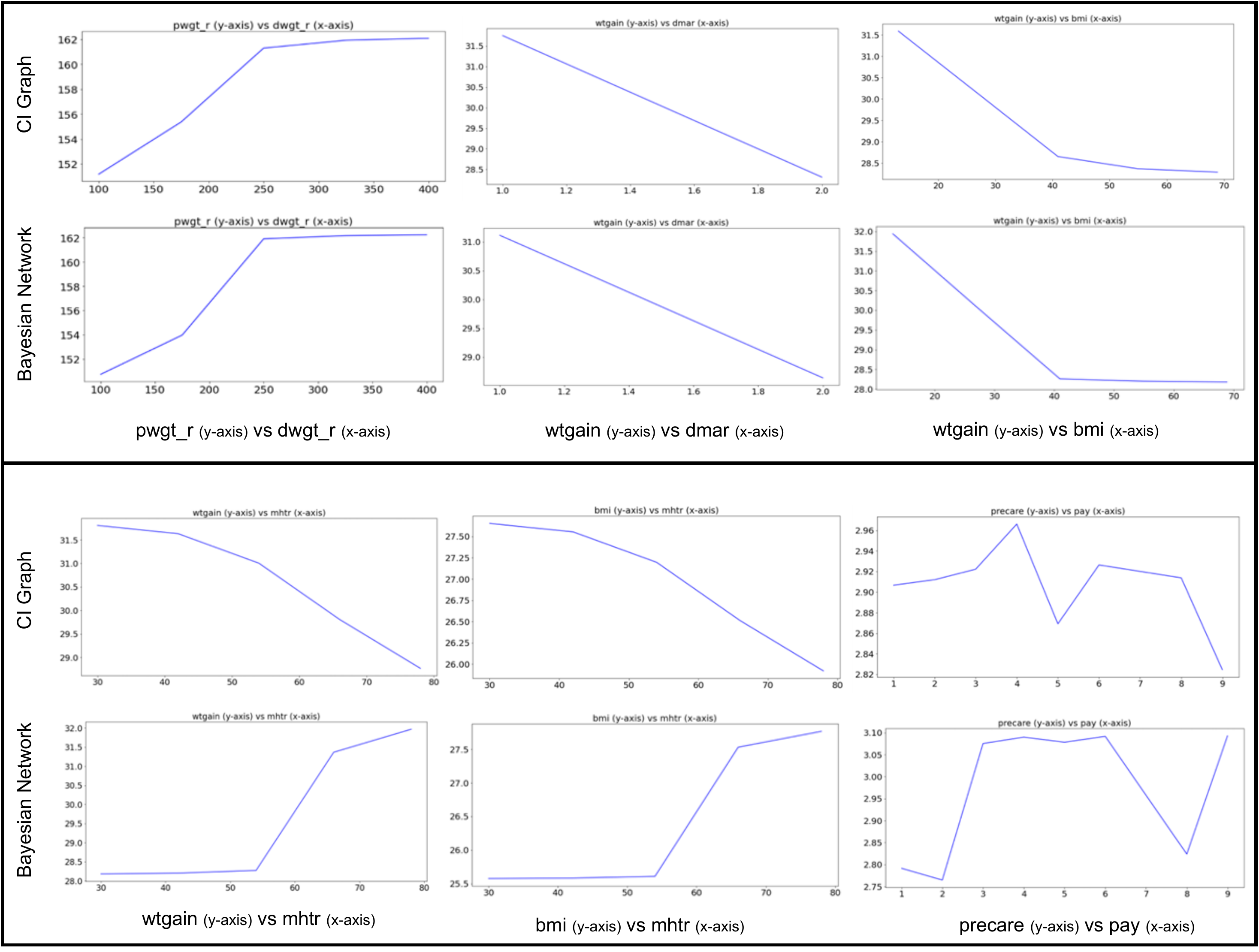}
\caption{\small \textbf{Evaluating effects of varying input graphs for learning \ngmsns.} Comparing the \ngm dependency plots recovered by using Bayesian Network graph vs the CI graph obtained by running \uglad. Similar architecture of \ngms were chosen and the data preprocessing was also kept as alike as possible. For the feature pairs in the top box, the trends match for both the graphs, while in the bottom box the dependency plots differ. We observed that the dependency trends discovered by the \ngm trained on the CI graph matches the correlation of the CI graph. Common edges present in both the graphs are: (pwgt-r, dwgt-r), (wtgain, mhtr), (bmi, mhtr), (precare, pay), edges only present in CI graph: (wtgain, dmar), (wtgain, bmi). It is interesting to observe that even for some common edges, eg. (wtgain, mhtr), that represents strong direct dependence between the features, the trends can still differ significantly. This highlights the importance of the input graph structure chosen to train \ngmsns. 
}
\label{fig:im2015-bn-vs-uglad-ngm-plots}
\end{figure*}

\textbf{Sensitivity to the input graph}: To study the effect of different graph structures on \texttt{NGMs}, we trained separate models on the Bayesian Network graph (moralized) and the CI graph from \uglad given in Fig.~\ref{fig:im2015-bn-graph} \& \ref{fig:im2015-ci-graph} respectively. 
%A detailed comparison of the two graphs is shown in Fig.~\ref{fig:im2015-compare-graph-bn-moral}.  
We plot the dependency functions between pairs of nodes based on the common and unique edges. For each pair of features, say $(f_1, f_2)$, the dependency function is obtained by running inference $P(f_1|f_2)$ by varying the value of $f_2$ over its range as shown in Fig.~\ref{fig:im2015-bn-vs-uglad-ngm-plots}. %in Appendix~\ref{apx:infant_mortality_extra}.
The two models largely agree on dependency patterns despite the differences between the two input graphs.  They also have similar prediction accuracy results as described in Section~\ref{sec:exp}.
%For more details and discussion, see Appendix~\ref{apx:infant_mortality_extra}.

\textbf{Dependency functions}: 
We plot the dependency functions between pairs of nodes based on the common and unique edges. For each pair of features, say $(f_1, f_2)$, the dependency function is obtained by running inference $P(f_1|f_2)$ by varying the value of $f_2$ over its range as shown in Fig.~\ref{fig:im2015-bn-vs-uglad-ngm-plots}.

Comparing \ngm inference in models trained with different input graphs (CI graph from \uglad and Bayesian network graph) shows some interesting patterns (see Fig.~\ref{fig:im2015-bn-vs-uglad-ngm-plots}):
\begin{itemize}[leftmargin=*,nolistsep]
\item Strong positive correlation of mother's delivery weight (\texttt{dwgt\_r}) with pre-pregnancy weight (\texttt{pwgt\_r}) is shown in both models.
\item Similarly, both models show that married mothers (\texttt{dmar}$=1$) are likely to gain more weight than unmarried (\texttt{dmar}$=2$).
\item Both models agree that women with high BMI tend to gain less weight during their pregnancies than women with low BMI.
\item A discrepancy appears in cases of the dependence of both BMI and weight gain during pregnancy on mother's height (\texttt{mhtr}).  According to the \ngm trained with a BN graph, higher weight gain and higher BMI are more likely for tall women, while the CI-trained \ngm shows the opposite. 
\item Possibly the most interesting are the graphs showing the dependence of the timing a women starts prenatal care (\texttt{precare} specifies the month of pregnancy when prenatal care starts) on the type of insurance she carries. For both models, Medicaid (1) and private insurance (2) mean early start of care and there is a sharp increase (delay in prenatal care start) for self-pay (3) and Indian Health Service (4). Models disagree to some extent on less common types of insurance (military, government, other, unknown).  

\end{itemize}

In general, the two models dependency functions agree to a larger extent than the dissimilarities between the two graphs would suggest.

\end{document}